%% file: main.tex
\documentclass{dhbenelux}

\usepackage{booktabs} % for toprule, midrule, bottomrule in tables
\usepackage{authblk} % for multiple authors and affiliations
\usepackage{graphicx} % to inlcude graphics with \includegraphics
\usepackage{tabularx}
\usepackage{dsfont}
\usepackage{bbold}
\usepackage{listings}
\usepackage{pifont}

\usepackage{multirow}% http://ctan.org/pkg/multirow

\usepackage{graphicx}
\newcommand*\rot[1]{\rotatebox{90}{#1}}

\usepackage{xcolor}
\usepackage{csvsimple}

\definecolor{codegreen}{rgb}{0,0.6,0}
\definecolor{codegray}{rgb}{0.5,0.5,0.5}
\definecolor{codepurple}{rgb}{0.58,0,0.82}
\definecolor{backcolour}{rgb}{0.95,0.95,0.92}

\lstdefinestyle{mystyle}{
    backgroundcolor=\color{backcolour},   
    commentstyle=\color{codegreen},
    keywordstyle=\color{magenta},
    numberstyle=\tiny\color{codegray},
    stringstyle=\color{codepurple},
    basicstyle=\ttfamily\footnotesize,
    breakatwhitespace=false,         
    breaklines=true,                 
    captionpos=b,                    
    keepspaces=true,                 
    numbers=left,                    
    numbersep=5pt,                  
    showspaces=false,                
    showstringspaces=false,
    showtabs=false,                  
    tabsize=2
}

\usepackage{enumerate}
\usepackage{amsmath}
\usepackage{txfonts}
\usepackage{amssymb}

\usepackage[normalem]{ulem} %sout barrer
\usepackage{enumitem}%label items

\usepackage{bbm}% mathbb avec des chiffres
\usepackage{dsfont} % mathbb pour les double barre sur un charactère
\usepackage{xspace} %espace si lettre, rien si ponctuation
\usepackage{relsize} %caractères plus grands

\usepackage{mdframed}
\input{Notations}

\newtheorem{mydef}{Definition}

\newtheorem{mypropo}{Proposition}

\newtheorem{mytheo}{Theorem}

\newtheorem{myfac}{Fact}
\newtheorem{mydata}{Framework}
\newtheorem{mymodel}{Model}

\newtheorem{const}{Construction}

\author[1]{Katia Meziani, Aminata Ndiaye, Benjamin Riu}

\title{%Chapter 5\\
\ourmeth : Marginal Contrastive Discrimination}

\begin{document}

\maketitle
% include copright statement on first page:
%\thispagestyle{papertitlepage} 

\begin{abstract}
    We consider the problem of conditional density estimation (\cde), which is a major topic of interest in the fields  of statistical and machine learning. Our method, called Marginal Contrastive Discrimination, \ourmeth, reformulates the conditional density function into two factors, the marginal density function of the target variable and a ratio of density functions which can be estimated through binary classification. Like noise-contrastive methods, \ourmeth can leverage \textit{state-of-the-art} supervised learning techniques to perform \cde, including neural networks. Our benchmark reveals that our method significantly outperforms in practice existing methods on most density models and regression datasets.
\end{abstract}

%%%%%%%%%%%%%%%%%%%%%%%%%%%%
\section{Introduction}
%%%%%%%%%%%%%%%%%%%%%%%%%%%%

\input{Introduction}

%%%%%%%%%%%%%%%%%%%%%%%%%%%%
\section{Marginal Contrastive Discrimination}\label{Sec:theory}
%%%%%%%%%%%%%%%%%%%%%%%%%%%%

%%%%%%%%%%%%%%%%%%%%%%%%%%%%
\subsection{Setting}
%%%%%%%%%%%%%%%%%%%%%%%%%%%%

In this \ourpaper, we consider three frameworks corresponding to three different situation in practice. 
 
 \begin{mdframed}
\begin{mydata}\label{DATA1}\textbf{[Independent Identically Distributed Samples]}\\
In this most classic setting, we consider $\mDxy = \{(\vX_i, \vY_i) \}_{i=1,\cdots, \SampSizexy}\,$ a training dataset of size $\SampSizexy \in \eN^*$ such that $\forall i = 1, \cdots, \SampSize$, the $(\vX_i, \vY_i)$ are $\pIID$ of density $\pf{\vX,\vY}$. 
\end{mydata}
\end{mdframed}

\medskip
In practice, it is often the case that additional observations are available but without the associated target values and vice versa (Framework~\ref{DATA2}). In this \ourpaper we show that it is possible to take advantage of these additional samples to increase the size of the training dataset without using any additional unsupervised learning techniques.\\

 \begin{mdframed}
\begin{mydata}\label{DATA2}\textbf{[Additional Marginal Data]}
\noindent\\
\noindent In this framework, we still consider a $\pIID$ training dataset of size $\SampSizexy \in \eN^*$ of density $\pf{\vX,\vY}$ denoted $\mDxy = \{(\vX_i, \vY_i) \}_{i=1,\cdots, \SampSizexy}$. Moreover we assume we have one or two additional datasets.
\begin{itemize}
    \item Let $\mDx = \{\vXtild_i \}_{i=1,\cdots, \SampSizex}$  be $\pIID$ an additional dataset of size $\SampSizex \in \eN$ of density  $\pf{\vX}$. 
    \item Let $\mDy = \{\vYtild_i \}_{i=1,\cdots, \SampSizey}$  be $\pIID$ an additional dataset of size $\SampSizey \in \eN$ of density  $\pf{\vY}$. 
\end{itemize}
We assume that $\mDxy$, $\mDx$ and $\mDy$ are independent.
\end{mydata}
\end{mdframed}

\medskip
Let us introduce a last framework, the one where more than one target value is associated to the same observation. To our knowledge, the article of (\cite{bott_nonparametric_2017}) is the only attempt to deal with this case. We show in this \ourpaper that our method can exploit and take advantage of these additional targets, again, without requiring an additional learning scheme. This can be the case for example in mechanics when performing fatigue analysis (\cite{bott_nonparametric_2017,manson_fatigue_1965}).\\

  \begin{mdframed}
\begin{mydata}\label{DATA3}\textbf{[Multiple Target per Sample]}\\
Let $(\vX, \mathbb{Y})$ be a couple of random variables taking values in $\pOmegX \times \pOmegY^\SampSizeM$ of density $\pf{\vX, \mathbb{Y}  }$. Let $\mDxyM = \{(\vX_i,\mathbb{Y}_i) \}_{i= 1}^{\SampSizexy}$ a training dataset of size $\SampSizexy \in \eN^*$ such that 
\begin{itemize}
    \item The  $\{\vX_i \}_{i = 1, \cdots, \SampSizexy}$ are sampled such that the $\vX_i$ are $\pIID$ of density $\pf{\vX}$.
    \item The $\{\mathbb{Y}_i \}_{i = 1, \cdots, \SampSizexy}$ are sampled such that the $\mathbb{Y}_i \pCond \vX_i$ are $\pIID$ of density $\pf{\mathbb{Y} \pCond \vX}$.
\end{itemize}
\end{mydata}
\end{mdframed}

%%%%%%%%%%%%%%%%%%%%%%%%%%%%%%%%%%%%%%%%%%%%%
 \subsection{Contrast function}
 %%%%%%%%%%%%%%%%%%%%%%%%%%%%%%%%%%%%%%%%%
 
 Our method, \ourmeth, is grounded on a trivial approached based on the successive application of the Bayes' formula \eqref{eq:Bayes0}. \\

We reformulate the problem differently from the existing noise contrastive methods, focusing on the contrast between the joint law $\pf{\vX,\vY}$ and the marginal laws $\pf{\vX}$ and $\pf{\vY}$. To do this, we define (Definition~\ref{def:MCF}) a new contrast $\contrast(\cdot,\cdot)$, called the marginal contrast function.

\begin{mdframed}
\begin{mydef}\label{def:MCF}\textbf{[Marginal Contrast function $\MCF(\ratio)$]}\\
Let $\ratio \in \pOmegr$ be a real number. Consider $(\vX,\vY)$ a couple of random variables taking values in $\pOmegX \times \pOmegY$. The Marginal Contrast Function with ratio $\ratio$ of the couple $(\vX,\vY)$, denoted $\contrast(\cdot,\cdot)$, is defined as:
\begin{align*}
\pOmegX \times \pOmegY &\eVers \pOmegq \\
(x, y) &\Associe \contrast(x,y) := \frac{\ratio \pf{\vX,\vY}(x,y)}{\ratio\pf{\vX,\vY}(x,y) + (1-\ratio)\pf{\vX}(x)\pf{\vY}(y)}.
\end{align*}
\end{mydef}
\end{mdframed}

\medskip
This new contrast is motivated by the Fact~\ref{fac:MCF}: \cde is equivalent to the marginal density of $\vY$ and the marginal contrast function $\contrast$. The \cde{} task is therefore reduced to the estimation of the constrast and a marginal density.\\

\begin{mdframed}
\begin{myfac}\label{fac:MCF}
Let $\ratio \in \pOmegr$ be a real number. Consider $(\vX,\vY)$ a couple of random variables taking values in $\pOmegX \times \pOmegY$. For all $(x, y) \in \pOmegX \times \pOmegY$, we have 
$$
\pf{\vY\pCond \vX = x}(y) = \pf{\vY}(y) \frac{\contrast(x,y)}{1-\contrast(x,y)} \frac{1-\ratio}{\ratio}
$$
where $\contrast(\cdot,\cdot)$ denotes the $\MCF(\ratio)$.
\end{myfac}
\end{mdframed}

\medskip

In the next section, we determine the conditions (Marginal Discrimination Conditions) under which the contrast function estimation can be transformed into an easy supervised learning estimation problem.

%%%%%%%%%%%%%%%%%%%%%%%%%%%%%%%%%%%%%%%%%%%%%%%%%%%%%%%%
 \subsection{Marginal Discrimination Conditions}
 %%%%%%%%%%%%%%%%%%%%%%%%%%%%%%%%%%%%%%%%%%%%%%%%%%%%%%%%

To transform the problem of estimating $\contrast$ into a problem of supervised learning, we first need to introduce a couple of random variables $(\vW, \vZ)$ satisfying the Marginal Discrimination Condition ($\MDC$) of the couple $(\vX,\vY)$ with ratio $\ratio\in(0,1)$.\\

\begin{mdframed}
\begin{mydef}\label{def:MDC}\textbf{[Marginal Discrimination Condition $\MDC(\ratio)$]}\\
Let $\ratio \in \pOmegr$ be a real number. Consider $(\vX,\vY)$ two random variables taking values in $\pOmegX \times \pOmegY$. A couple of random variables $(\vW, \vZ)$ is said to satisfy the Marginal Discrimination Condition ($\MDC$) of the couple $(\vX,\vY)$ with ratio $\ratio$ if 
\begin{enumerate}[label=(\textbf{Cd~\arabic*})]
    \item\label{MDC1} The random variable $\vZ$ follows a Bernoulli law of parameter $\ratio$ ($\vZ \sim \pBB(r)$).
    \item \label{MDC2} The support of $\vW$ is $\,\mathcal{\vW} = \mathcal{\vX} \times \mathcal{\vY}$.
    \item \label{MDC3} For all $
(x,y) \in \mathcal{X} \times \mathcal{\vY},$ we have $\pf{\vW }(x,y) = \ratio\pf{\vX,\vY}(x,y)+ (1-\ratio)\pf{\vX}(x)\pf{\vY}(y).
$
    \item \label{MDC4} For all $
(x,y,z) \in \pOmegX \times \pOmegY \times \pOmegZ$, we have
$$ \pf{\vW \pCond \vZ=z}(x,y) = \Indi_{z = 1}\pf{\vX,\vY}(x,y)+ \Indi_{z = 0}\pf{\vX}(x)\pf{\vY}(y).
$$
\end{enumerate}
\end{mydef}
\end{mdframed}

Remark that condition \ref{MDC3} is satisfied if conditions \ref{MDC1},\ref{MDC2} and \ref{MDC4} are verified. These conditions are sufficient to characterise both the joint and marginal laws of the couple $(\vW, \vZ)$. 

%%%%%%%%%%%%%%%%%%%%%%%%%%%%%%%%%%%%%%%%%%%%%%%%%%%%%%%%%%%%%%%
\paragraph{Estimation of the contrast function through supervised learning}

The Proposition~\ref{pro:equiv} specifies how the marginal contrast function $\contrast$ problem can be estimated by a supervised learning task, using either a regressor or a binary classifier, provided that we have access to a sample of identically distributed ($\pUID$) of random variables that satisfies the $\MDC(\ratio)$.\\

\begin{mdframed}
\begin{mypropo}\label{pro:equiv}\textbf{[Constrast Estimation]}\\
Let $\ratio \in \pOmegr$ be a real number. Consider $(\vX,\vY)$  a couple of random variables taking value in $\pOmegX \times \pOmegY$. For all $(x, y) \in \pOmegX \times \pOmegY$, the Marginal Contrast Function of couple $(\vX,\vY)$ with ratio $r$ denoted by $\contrast$
satisfies the following property:

$$
\contrast(x,y) = \pE[\vZ \pCond \vW = (x,y)] = \pP[\vZ = 1 \pCond \vW = (x,y)].
$$
\end{mypropo}
\end{mdframed}

The proof is given in section~\ref{sec:Proofequiv}. It remains to prove that it is possible to construct a training set of identically distributed $(\pUID$) samples of $(\vW, \vZ)$ using the elements of the original dataset.
%We then need to prove that it is possible to build a training set of $\pIID$ samples of $(\vW, \vZ)$ using the elements of the original dataset.

%%%%%%%%%%%%%%%%%%%%%%%%%%%%%%%%%%%%%%%%%%%%%%%%%%
\section{Contrast datasets construction}
%%%%%%%%%%%%%%%%%%%%%%%%%%%%%%%%%%%%%%%%%%%%%%%%%%

%%%%%%%%%%%%%%%%%%%%%%%%%%%%%%%%%%%%%%%%%%%%
\subsection{Classical Dataset (Framework~\ref{DATA1})}
%%%%%%%%%%%%%%%%%%%%%%%%%%%%%%%%%%%%%%%%%%%%

Theorem~\ref{pro:ConsIID} establishes the existence of an $\pIID$ sample of $(\vW, \vZ)$ satisfying the $\MDC(\ratio)$ in Framework~\ref{DATA1}. In its proof (section~\ref{Sec:ProofConsIID}), such of construction based on the original data set is derived. From now and for all $\alpha \in \eR$, the quantity $\ArrB{\alpha}$ denote the largest integer value smaller or equal to $\alpha$.

\begin{mdframed}
\begin{mytheo}\textbf{[Construction of an $\pIID$ training Set]}\\
\label{pro:ConsIID}
Let $\ratio \in \pOmegr$ be a real number. Consider the dataset $\mDxy$ defined in Framework~\ref{DATA1}.\\

\noindent Then, we can construct a dataset $\mDwz{\SampSizewz} = \{(\vW_i, \vZ_i) \}_{i=1,\cdots, \SampSizewz}\,$ of size $\SampSizewz = \ArrB{\SampSize/2}$ of $\pIID$ observations %such that,  $\forall i=1,\cdots, \SampSizewz, (\vW_i, \vZ_i)$ 
satisfying the $\MDC(\ratio)$.
\end{mytheo}
\end{mdframed}

\medskip
Note that, in practice, having access to a larger data set improves the results considerably. By dispensing with the independence property, it is possible to construct a much larger data set without deteriorating the results (see numerical experiments). This is the purpose of Theorem~\ref{pro:ConsUID}\\

\begin{mdframed}
\begin{mytheo}\label{pro:ConsUID}\textbf{[Construction of a larger $\pUID$ training Set]}\\
Consider the dataset $\mDxy$ defined in Framework~\ref{DATA1}.
%Consider the framework~\ref{DATA1}.
Moreover, assume that $\mDxy$ is such that $\forall (i,j) \in \{1,\cdots,\SampSizexy\}^2$  with $i \neq j$
$$
\vX_i \neq \vX_j \  and \ \vY_i \neq \vY_j
$$
\noindent
%Then, for any couple of integers $(n_J,n_M)$ such that $1 \leq n_J \leq \SampSizexy$, $1 \leq n_M \leq \SampSizexy(\SampSizexy - 1)$,
Then, for any couple of integers $(n_J,n_M)$ such that
$$\left\{
\begin{array}{l}
  1 \leq n_J \leq \SampSizexy\\
1 \leq n_M \leq \SampSizexy(\SampSizexy - 1)
\end{array}
\right.
$$
we can construct a dataset $\mDwz{\SampSizewz} = \{(\vW_i, \vZ_i) \}_{i=1,\cdots, \SampSizewz}$ of size $\SampSizewz = n_J + n_M$ of $\pUID$ random %variables %such that $\forall i=1,\cdots, \SampSizewz, \, (\vW_i, \vZ_i)$ satisfies the $\MDC\left(\frac{n_J}{\SampSizewz}\right)$.
observations %such that,  $\forall i=1,\cdots, \SampSizewz, (\vW_i, \vZ_i)$ 
satisfying the $\MDC\left(\frac{n_J}{\SampSizewz}\right)$.
\end{mytheo}
\end{mdframed}

\medskip

Note first that we can at most construct a dataset of size  $\SampSizewz = \SampSize^2$, with $r = \frac{1}{\SampSizexy}$. On the other hand, if we want to have $r = \frac{1}{2}$, we can generate a dataset of size $\SampSizewz = 2\SampSizexy$. 
Second, the additional conditions on the dataset are introduced to exclude the trivial case where a larger dataset is constructed by simply repeating the existing samples. In practice, we can always avoid this case by removing redundant samples. Note, however, that the repetition of values occurs with probability $0$, almost surely, because we consider continuous densities. 
Finally, the complete construction of such a dataset is described in section~\ref{Sec:ProofConsUID}

%%%%%%%%%%%%%%%%%%%%%%%%%%%%%%%%%%%%%%%%%%%%
\subsection{Additional Marginal Data (Framework~\ref{DATA2})}
%%%%%%%%%%%%%%%%%%%%%%%%%%%%%%%%%%%%%%%%%%%%

Now consider that we have additional features and/or targets for which the target or associated feature is not available. We include this additional data in our training process without using a semi-supervised scheme.\\

\begin{mdframed}
\begin{mytheo}\label{pro:ConsIIDAM}\textbf{[Construction of an $\pIID$ training set]}\\
Let $\ratio \in \pOmegr$ be a real number. Consider the datasets  defined in Framework~\ref{DATA2}. Set 
$$\SampSizewz = min\left(\SampSizexy, \ArrB{\frac{\SampSize + \SampSizex + \SampSizey}{2}}\right),
$$
\noindent then, we can construct $\mDwz{\SampSizewz} = \{(\vW_i, \vZ_i) \}_{i=1,\cdots, \SampSizewz}$ a dataset of size  $\SampSizewz$ of $\pIID$ observations satisfying the $\MDC(\ratio)$.

%Let $\SampSizewz = min\left(\SampSizexy, \ArrB{\frac{\SampSize + \SampSizex + \SampSizey}{2}}\right)$, Then we can construct $\mDwz{\SampSizewz} = \{(\vW_i, \vZ_i) \}_{i=1,\cdots, \SampSizewz}$ a dataset of size  $\SampSizewz$ of $\pIID$ observations, such that $(\vW_i, \vZ_i)$ satisfies the $\MDC(\ratio)$.
\end{mytheo}
\end{mdframed}

\medskip
This theorem implies that as soon as we have $\SampSizex + \SampSizey \geq \SampSize$, we can generate a training set for the discriminator as large as the original set, i.e. $\SampSizewz = \SampSizexy$. In practice, this can happen in many cases. For example, when data annotation is expensive or difficult, we often have $\SampSizex >> \SampSizexy$. At the same time, to use contrastive marginal discrimination to estimate the conditional density, we need to know or estimate the marginal density $\pf{\vY}$. The proof of this theorem is done in section~\ref{Sec:ProofConsIIDAM}\\

\begin{mdframed}
\begin{mytheo}\label{pro:ConsUIDAM}\textbf{[Construction of a larger  $\pUID$ training set]}\\
Consider the dataset  defined in Framework~\ref{DATA3}. Moreover assume \\
%Consider the framework~\ref{DATA2} and assume that:\\

\noindent$\bullet$ The dataset $\mDxy$ is such that $\forall (i,j) \in \{1,\cdots,\SampSizexy\}^2$ $s.t.$ $i \neq j$
$$
\vX_i \neq \vX_j  \  and  \ \vY_i \neq \vY_j.
$$
$\bullet$ The datasets $\mDx{\SampSizex}$ and $\mDy{\SampSizey}$ are $s.t.$ $\forall (i,j) \in\{1,\cdots,\SampSizex\}^2$ and $\forall (i',j') \in \{1,\cdots,\SampSizey\}^2$
$$
\vXtild_i \neq \vXtild_j  \  and  \  \vYtild_{i'} \neq \vYtild_{j'}
$$
%$\bullet$ Assume that $\mDx$ and $\mDy$ are such that $\forall (i,j) \in \llbracket 1,\SampSizexy \rrbracket^2$  $i \neq j:\, \vX_i \neq \vX_j,$ and $ \vY_i \neq \vY_j$.\\
%Also assume that $\mDx{\SampSizex}$ and $\mDy{\SampSizey}$ such that 
%$$\left\{
%\begin{array}{ll}
 % \vXtild_i \neq \vXtild_j \,\, \forall (i,j) \in\{1,\cdots,\SampSizex\}^2,\\
%  \vYtild_i \neq \vYtild_j \,\, \forall (i,j) \in \{1,\cdots,\SampSizey\}^2.
%\end{array}
%\right.$$
$\bullet$ Moreover, we assume that $\mDxy$, $\mDx$ and $\mDy$ are such that $\forall (i,i') \in \{1,\cdots,\SampSizexy\}\times \{1,\cdots,\SampSizex\}$ and $\forall (j,j') \in \{1,\cdots,\SampSizexy\}\times \{1,\cdots,\SampSizey\}$
$$
\vX_i \neq \vXtild_{i'}  \  and  \  \vY_j \neq \vYtild_{j'}
$$
%$$\left\{
%\begin{array}{ll}
%  \vX_i \neq \vXtild_{i'}\,\, \forall (i,i') \in \{1,\cdots,\SampSizexy\}\times \{1,\cdots,\SampSizex\},\\
%  \vY_j \neq \vYtild_{j'} \,\, \forall (j,j') \in \{1,\cdots,\SampSizexy\}\times \{1,\cdots,\SampSizey\}.
%\end{array}
%\right.$$
Then, for any couple of integers $(n_J,n_M)$ such that
$$\left\{
\begin{array}{l}
  1 \leq n_J \leq \SampSizexy\\
1 \leq n_M \leq (\SampSizexy + \SampSizex)(\SampSizexy+\SampSizey) - \SampSizexy
\end{array}
\right.$$
we can a dataset $\mDwz{\SampSizewz} = \{(\vW_i, \vZ_i) \}_{i=1,\cdots, \SampSizewz}$ of size $\SampSizewz = n_J + n_M$ of $\pUID$ random observations satisfying the $\MDC(\frac{n_J}{\SampSizewz})$.
%$1 \leq n_J \leq \SampSizexy$ and \linebreak$1 \leq n_M \leq (\SampSizexy + \SampSizex)(\SampSizexy+\SampSizey) - \SampSizexy$, 
%we can construct a dataset $\mDwz{\SampSizewz} = \{(\vW_i, \vZ_i) \}_{i=1,\cdots, \SampSizewz}$ of size $\SampSizewz = n_J + n_M$ of identically distributed random variables such that \linebreak$\forall i=1,\cdots, \SampSizewz, \, (\vW_i, \vZ_i)$ satisfies the $\MDC(\frac{n_J}{\SampSizewz})$. 
\end{mytheo}
\end{mdframed}

\noindent Here again, it is possible to build a much larger $\pUID$ training dataset under some weak assumption. Indeed, the repetition of values occurs with probability $0$, almost surely. The complete construction of such a dataset is described in section~\ref{Sec:ProofConsUIDAM}
\medskip

%%%%%%%%%%%%%%%%%%%%%%%%%%%%%%%%%%%%%%%%%%%%%%%%%%%%
\subsubsection{Multiple targets per observations (Framework~\ref{DATA3})}
%%%%%%%%%%%%%%%%%%%%%%%%%%%%%%%%%%%%%%%%%%%%%%%%

In Framework~\ref{DATA3}, to each of  the $\SampSizexy$ observations $\vX_i$, there exists a $\SampSizeM$-associated target $\mathbb{Y}_i=\big(\vY_{i,1},\cdots,\vY_{i,\SampSizeM}\big)$ of $\pIID$ components such that the $\left(\vY_{i,j}\pCond \vX_i\right)_{j=1,\cdots,\SampSizeM}$ are $\pIID$.  In this setting, it is still possible to construct, under some weak assumption, a larger $\pUID$ training set satisfying the $\MDC$. Recall, the repetition of values occurs with probability $0$, almost surely. Note that we can construct, at most, a data set of size $\SampSizexy^2 \times \SampSizeM$, with $\ratio = \frac{1}{\SampSize}$. On the other hand, if we want to have a ratio of $\ratio = \frac{1}{2}$, the generated dataset will be of size $2\SampSizexy \times \SampSizeM$.\\

\begin{mdframed}
\begin{mytheo}\label{pro:ConsUIDMT}\textbf{[Construction of a larger $\pUID$ training set]}\\
\noindent Consider the dataset $\mDxyM = \{(\vX_i,\mathbb{Y}_i) \}_{i= 1}^{\SampSizexy}$ a training dataset of size $\SampSizexy \in \eN^*$ defined in Framework~\ref{DATA3}. Assume that \\
$\bullet$ For all $(i,i') \in \{1,\cdots,\SampSizexy\}^2$ such that $i \neq i'$
$$
\vX_i \neq \vX_{i'}.
$$
$\bullet$ For all $(i,j),(i',j') \in \{1,\cdots,\SampSizexy\}\times \{1,\cdots,\SampSizeM\}$ such that $i \neq i'$ or $j \neq j'$
$$
\vY_{i,j} \neq \vY_{i',j'}
$$
\noindent Then, for any couple of integers $(n_J,n_M)$ such that
$$\left\{
\begin{array}{l}
  1 \leq n_J \leq \SampSizexy \times \SampSizeM\\
1 \leq n_M \leq \SampSizexy (\SampSizexy - 1) \times \SampSizeM
\end{array}
\right.$$
we can construct a dataset $\mDwz{\SampSizewz} = \{(\vW_i, \vZ_i) \}_{i=1,\cdots, \SampSizewz}$ of size $\SampSizewz = n_J + n_M$ of $\pUID$ random observations satisfying the $\MDC\left(\frac{n_J}{\SampSizewz}\right)$.
\end{mytheo}
\end{mdframed}

The complete construction of such a dataset is described in section~\ref{Sec:ProofConsUIDMT}

\input{Experiments}
\input{Proofs}
\section{Conclusion}
%%%%%%%%%%%%%%%%%%%%%%%%%%%%%%%%%%%%%%%%%%%%%%%%
In this \ourpaper, we consider the problem of conditional density estimation. We introduce a new method, $\mcd$ inspired by contrastive learning. $\mcd$ reformulates the initial task into a problem of supervised learning. We present construction techniques to produce contrast dataset of \pIID or \pUID samples with far more observations than in the original dataset. We also provided construction techniques to take advantage of unlabeled observations and more than one target value per observation. We evaluate our method on a benchmark of both density models and real-world datasets, and obtain excellent results in most cases, especially when $\mcd$ is combined with Neural Networks.\\

There are still many questions left open with regard to the appropriate choice of discriminator and construction strategy, notably the ratio $\ratio$. Besides, assessing the performances of \mcd on down-stream tasks such as quantile regression, variance estimation or outlier detection is also a promising future avenue of research.

\bibliography{reference}
\input{Appendix}
\end{document}

%% file: Notations.tex
\newcommand{\ourpaper}{article\,}
\newcommand{\ourmeth}{\textbf{MCD}\,}
\newcommand{\mcd}{\textbf{MCD}\,}
\newcommand{\MCF}{\ensuremath{\textbf{MCF}}}
\newcommand{\MDC}{\ensuremath{\text{MDcond}}}
\newcommand{\cde}{\textbf{CDE}\,}

\newcommand{\NeurN}{\textbf{NN}\,}

\newcommand{\rescale}{MCD.\textit{rescale}\,(pdf)}
\newcommand{\flex}{\textbf{FlexCode}\,}
\newcommand{\nnk}{\textbf{NNKCDE}\,}
\newcommand{\rfcde}{\textbf{RFCDE}\,}
\newcommand{\mdn}{\textbf{MDN}\,}
\newcommand{\kmn}{\textbf{KMN}\,}
\newcommand{\lsc}{\textbf{LSCond}\,}
\newcommand{\dcde}{\textbf{Deepcde}\,}
\newcommand{\nf}{\textbf{N.Flow}\,}
\newcommand{\CAT}{\textbf{CatBoost}\,}
\newcommand{\XGB}{\textbf{XGboost}\,}
\newcommand{\LGBM}{\textbf{LGBM}\,}
\newcommand{\Enet}{\textbf{E.Net}\,}
\newcommand{\RF}{\textbf{RF}\,}
\newcommand{\XRF}{\textbf{XRF}\,}
\newcommand{\SNN}{\textbf{SNN}\,}

\newcommand{\GLU}{\textbf{GLU}}
\newcommand{\ResBlock}{\textbf{ResBlock}}
\newcommand{\MLP}{\textbf{MLP}\,}
\newcommand{\NoDO}{\textbf{no-D.O.}}

\newcommand{\FLT}{\textit{freelunchtheorem}\,}
\newcommand{\packpython}{\textit{python}\,}
\newcommand{\packpytorch}{\textit{pytorch}\,}
\newcommand{\packtensorflow}{\textit{tensorflow}\,}
\newcommand{\packgithub}{\textit{github}\,}
\newcommand{\packCDEtools}{\textit{CDEtools}\,}
\newcommand{\packfreelunchtheorem}{\textit{freelunchtheorem}\,}

\newcommand{\dEcon}{\textbf{EconDensity}\,}
\newcommand{\dArma}{\textbf{ARMAJump}\,}
\newcommand{\dJump}{\textbf{JumpDiffusion}\,}
\newcommand{\dLinG}{\textbf{LinearGauss}\,}
\newcommand{\dStud}{\textbf{LinearStudentT}\,}
\newcommand{\dSkew}{\textbf{SkewNormal}\,}
\newcommand{\dBasic}{\textbf{BasicLinear}\,}
\newcommand{\dAsym}{\textbf{AsymmetricLinear}\,}
\newcommand{\dGMixt}{\textbf{GaussianMixt}\,}

\newcommand{\STEPUIDOne}{\textbf{Step (1)}\,}
\newcommand{\STEPUIDOneb}{\textbf{Step (1-bis)}\,}
\newcommand{\STEPUIDOnet}{\textbf{Step (1-ter)}\,}
\newcommand{\STEPUIDTwo}{\textbf{Step (2)}\,}
\newcommand{\STEPUIDThree}{\textbf{Step (3)}\,}
\newcommand{\STEPUIDFour}{\textbf{Step (4)}\,}
\newcommand{\STEPUIDFive}{\textbf{Step (5)}\,}

\newcommand{\SUBSTEPA}{\textbf{(A)}\,}
\newcommand{\SUBSTEPB}{\textbf{(B)}\,}
\newcommand{\SUBSTEPC}{\textbf{(C)}\,}
\newcommand{\SUBSTEPD}{\textbf{(D)}\,}
\newcommand{\SUBSTEPE}{\textbf{(E)}\,}

\newcommand{\STEPUIDNext}{\textbf{(Next-Step)}\,}

\newcommand{\eN}{\ensuremath{\mathbbm{N}}}

\newcommand{\eR}{\ensuremath{\mathbbm{R}}}

\newcommand{\eVers}{\ensuremath{\longrightarrow}}

\newcommand{\pE}{\ensuremath{\mathbb{E}}}
\newcommand{\pEz}{\ensuremath{\mathbb{E}_{Z}}}
\newcommand{\pP}{\ensuremath{\mathbb{P}}}

\newcommand{\pIID}{\ensuremath{i.i.d.}}
\newcommand{\pUID}{\ensuremath{i.d.}}

\newcommand{\pIndep}{\perp \!\!\! \perp}

\newcommand{\pOmegX}{\ensuremath{\mathcal{X}}}
\newcommand{\pOmegY}{\ensuremath{\mathcal{Y}}}

\newcommand{\pOmegZ}{\ensuremath{\{0;1\}}}
\newcommand{\pOmegr}{\ensuremath{(0,1)}}
\newcommand{\pOmegq}{\ensuremath{[0,1)}}

\newcommand{\pequal}{\ensuremath{{\buildrel \mathcal{D} \over =}}}

\newcommand{\pf}[1]{\ensuremath{\,\mathlarger{\textit{f}}_{#1}\,}} % densité
\newcommand{\pg}[1]{\ensuremath{\,\mathlarger{\textit{g}}_{#1}\,}} % autre densité

\newcommand{\pfhat}[1]{\ensuremath{\,\mathlarger{\widehat{\textit{f}}}_{#1}\,}} % densité
\newcommand{\pBB}{\ensuremath{\mathcal{B}}}% Bernouilli
\newcommand{\pUU}{\ensuremath{\mathcal{U}}}% Uniform
\newcommand{\pNN}{\ensuremath{\mathcal{N}}}% Bernouilli
\newcommand{\pCond}{\ensuremath{\,|\,}}% Conditionnellement à 
\newcommand{\ratio}{\ensuremath{r}}% Bernouilli

\newcommand{\pKL}{\textbf{KL}\,}
\newcommand{\pNLL}{\text{NLL}\,}

\newcommand{\Id}{\ensuremath{\mathbb{I}}}
\newcommand{\Zeros}{\ensuremath{\boldsymbol{0}}}
\newcommand{\transp}{\top}
\newcommand{\Indi}{\ensuremath{\mathds{1}}}
\newcommand{\Integ}[2]{\ensuremath{\displaystyle{\int}_{\!\!\!#1}^{\,#2}\!\!\!}}
\newcommand{\Associe}{\ensuremath{\longmapsto}}

\newcommand{\ArrB}[1]{\left\lfloor #1 \right\rfloor}

\newcommand{\SampSize}{\ensuremath{n}}
\newcommand{\SampSizexy}{\ensuremath{n}}
\newcommand{\SampSizeM}{\ensuremath{m}}
\newcommand{\SampSizex}{\ensuremath{n_x}}
\newcommand{\SampSizey}{\ensuremath{n_y}}
\newcommand{\SampSizewz}{\ensuremath{N}}

\newcommand{\SampSizewzx}{\ensuremath{N}_X}
\newcommand{\SampSizewzy}{\ensuremath{N}_Y}
\newcommand{\SampSizewzxy}{\ensuremath{N}_{X,Y}}
\newcommand{\DimInput}{\ensuremath{p}}
\newcommand{\DimOutput}{\ensuremath{k}}

\newcommand{\contrast}{q}
\newcommand{\contrasthat}{\widehat{q}}
\newcommand{\contrasttild}{\widetilde{q}}

\newcommand{\vX}{X}
\newcommand{\vXtild}{\widetilde{X}}

\newcommand{\vY}{Y}
\newcommand{\vYtild}{\widetilde{Y}}
\newcommand{\ytild}{\widetilde{y}}
\newcommand{\vZ}{Z}
\newcommand{\vW}{W}
\newcommand{\vWtild}{\widetilde{W}}
\newcommand{\vZtild}{\widetilde{Z}}

\newcommand{\mDxy}{\ensuremath{\mathcal{D}^{X,Y}_n}}
\newcommand{\mDxyM}{\ensuremath{\mathcal{D}_{n,m}^{X,\mathbb{Y}}}}
\newcommand{\mDwz}[1]{\ensuremath{\mathcal{D}}_{#1}^{W,Z}}

\newcommand{\mDx}{\ensuremath{\mathcal{D}_{n_x}^{X}}}
\newcommand{\mDy}{\ensuremath{\mathcal{D}_{n_y}^{Y}}}

\newcommand{\mS}{\ensuremath{\mathcal{S}}}
\newcommand{\mMx}{\ensuremath{\mathcal{S}^{\widetilde{X}}}}
\newcommand{\mMy}{\ensuremath{\mathcal{S}^{\widetilde{Y}}}}
\newcommand{\mMxy}{\ensuremath{\mathcal{S}^{\widetilde{X},\widetilde{Y}}}}

\newcommand{\mDxyRes}[1]{\ensuremath{\mathcal{D}^{X,Y}_{|#1}}}

\newcommand{\mDwzRes}[1]{\ensuremath{\mathcal{D}^{W,Z}_{|#1}}}

\newcommand{\mDxRes}[1]{\ensuremath{\mathcal{D}^{X}_{|#1}}}
\newcommand{\mDyRes}[1]{\ensuremath{\mathcal{D}^{Y}_{|#1}}}

\newcommand{\Tau}{\ensuremath{\mathcal{O}}}

%% file: Introduction.tex
We consider the problem of conditional density estimation (\cde), which is a major topic of interest in the fields  of statistical and machine learning.\\

We consider a couple of random variables $(\vX,\vY)$ taking values in $\pOmegX \times \pOmegY $ such that $\pOmegX \subseteq \eR^\DimInput$ and $\pOmegY \subseteq \eR^\DimOutput$. We assume in this \ourpaper that all random variables admit a density function with respect to a dominant measure. We also assume all these densities are proper $i.e.$ they integrate to 1. We denote $\pf{\vX}$ and $\pf{\vY}$ the marginal densities of $\vX$ and $\vY$ with respect to a dominant measure.
Our goal is to estimate the conditional density function:
\begin{align*}
\pOmegX \times \pOmegY &\eVers \eR_+ \\
(x, y) &\Associe \pf{\vY \pCond \vX = x}(y)
\end{align*}

This problem is at the root of the majority of machine learning tasks, including supervised and unsupervised learning or generative modelling. Supervised learning techniques aims at estimating the conditional mean. Meanwhile, in the binary classification setting, these two tasks are equivalent, since $\pE[Y\pCond X=x] = \pP[\vY=1\pCond \vX=x]$ . In the regression setting where $Y$ is a continuous variable, the conditional density is far more informative than the mean value. This is especially true when the conditional distribution is multi-modal, heteroscedastic  or heavy tailed.
Moreover, in many fields such as actuarial science, asset management,
climatology, econometrics, medicine or astronomy, one is interested in quantities other than expectation, such as higher order moments (variance, skewness
kurtosis), prediction intervals, quantile regression, outlier boundaries, etc.\\

Meanwhile, most unsupervised learning techniques aim to discover relationships and patterns between random variables. This corresponds to the joint density probability function estimation subtask, itself a subtask of \cde. Similarly, the field of generative modeling, whose goal is to generate synthetic data by expressing the joint distribution as a product of univariate conditional distributions with respect to latent variables, can also be considered a \cde subtask, as realistic images or sounds correspond to the modes of the distribution.\\

%%%%%%%%%%%%%%%%%%%%%%%%%%%%
\subsection{Related work}
%%%%%%%%%%%%%%%%%%%%%%%%%%%%
Historically, the first attempts were based on the use of Bayes' formula~\eqref{eq:Bayes0} which transforms the \cde into the estimation of two density functions, thus allowing the use of techniques dedicated to density estimation.
\begin{align}
\label{eq:Bayes0}
  \pf{Y\pCond X =x}(y) = \frac{\pf{X,Y}(x,y)}{\pf{X}(x)}, \, \pf{X}(x) \neq 0 
\end{align}

In this \ourpaper, we address the many real-world applications of supervised learning regressions where we have a one-dimensional target but a large number of features. A major flaw of reformulating a conditional density as the ratio of two densities, is that even if $Y$ is low dimensional, we will still incur the curse of dimensionality if $X$ is high-dimensional. Although techniques exist to alleviate the curse to some extent by leveraging sparsity or other properties of $\vX$, it would be far easier to reformulate  the \cde so that only the estimate of the marginal density $\pf{\vY}$ is required instead of $\pf{\vX}$.

%%%%%%%%%%%%%%%%%%%%%%%%
\paragraph{Nonparametric density estimation.}
Nonparametric estimation is a powerful tool for density estimation as it does not require any prior knowledge of the underlying density. One of the first intuitions of kernel estimators was proposed by \cite{rosenblatt_remarks_1956} and later by \cite{parzen_estimation_1962}. Kernel estimators were  then widely studied ranging from bandwidth selection (\cite{goldenshluger_bandwidth_2011}), non linear aggregation (\cite{rigollet_linear_2006}), computational optimisation (\cite{langrene_fast_2020}) to the extension to conditional densities (\cite{bertin_adaptive_2014}). We refer to  \cite{silverman_density_2017} and references therein for interested reader. A very popular and effective nonparametric method is the $k$-Nearest Neighbors (\cite{fix_discriminatory_1989}), but like other nonparametric methods (kernel estimators, histogram \cite{pearson_contributions_1895},...), a main limitation  is the curse of dimensionality (\cite{scott_feasibility_1991,nagler_evading_2016}). \cite{silverman_density_2017} showed that, in a density estimation setting, the number of points $n$, needed to obtain equivalent results when fitting a $d$-dimensional random variable, grows at a rate of $n^{\frac{4}{4 + d}}$. Meanwhile, the impact of dimensionality on the computational time also scales exponentially. To the best of our knowledge, the best performing method for kernel estimation, based on a divide-and-conquer algorithm (\cite{langrene_fast_2020}), has a complexity of $\Tau (n \log(n)^{\max(d-1,1)})$. Both of these relationships are compounded, as  higher dimensionality means that exponentially more data is required and the computational time relative to the size of the dataset will also grow exponentially.\\
%In this Chapter, we implement different kernel estimators in python with \textit{statsmodels} (\cite{noauthor_kernel_nodate-1}).}

%%%%%%%%%%%%%%%%%%%%%%%%%%
\paragraph{Noise Contrastive methods.}

Another important family of techniques for density estimation is noise-contrastive learning (\cite{gutmann_noise-contrastive_2010}). These techniques reformulate the estimation of the density into a binary classification problem (up to a constant). It consists in introducing a known  probability density $g(\cdot)$ and sampling from it a synthetic dataset. The latter is concatenated with the original dataset, then a target value $\vZ_i$ is associated such that $\vZ_i$ is equal to $1$ if the observation comes from the original dataset and $0$ otherwise.
A \textbf{contrast} $\contrast(\cdot)$, which can be, under certain conditions on $g$, directly related to the density of the original observations is introduced to obtain an estimation of the density function :
\begin{equation}
\label{eq:noisecontrastive}
\contrast (x):= \pP[\vZ=1\pCond \vX = x]= \frac{\pf{\vX}(x)}{\pf{\vX}(x)+g(x)}\quad \text{and}\quad \pf{\vX}(x) = g(x)\frac{\contrast(x)}{1-\contrast(x)}.
\end{equation}
A binary classifier is then trained to predict this target value conditionally on the associated observation.\\

Contrast learning has been successfully applied in the area of self-supervision learning (\cite{jaiswal_survey_2020,he_momentum_2020}), especially  on computer vision tasks (\cite{bachman_learning_2019}).
Many extensions and improvements have been made, including w.r.t. the learning loss function (\cite{khosla_supervised_2021}), data augmentation techniques  (\cite{chen_simple_2020}), network architectures (\cite{he_momentum_2020}) and computational efficiency (\cite{yeh_decoupled_2021}). Our proposed method is partially based on this technique, with two major differences. First, our technique addresses \cde problem and not density estimation or self-supervised learning. Second, the distribution we choose is unknown, potentially intractable, and/or with a large set of highly dependent components, which violates the usual restrictions for performing noise contrastive density estimation  but allows us to taylor the noise distribution precisely to the estimated distribution.
This allows us to fit the noise distribution precisely to the estimated distribution. To the best of our knowledge, the technique closest to our method is designed to evaluate the deviation from the independence setting, $i.e.$ when all random features $\{\vX_j\}_{j=1,\cdots,\DimInput}$ are independent, in an unsupervised framework. It has been briefly described in the second edition of \cite{tibshirani_elements_nodate} (pages 495-497) where, based on the noise contrastive reformulation given above with $g(\cdot) = \Pi_{j=1}^{\DimInput} \pf{\vX_j}(\cdot)$.  Note that $g$ is unknown and corresponds to what we will call later the noise distribution. Nevertheless, noise samples can be generated by applying a random permutation on the feature columns of the dataset, which is sufficient to discover association rules between the $\vX$ features but not to estimate the density function $\pf{\vX}$. On the other hand, although each component of $\vX$ is one-dimensional by definition, the errors made when estimating the marginal densities $\DimInput$ will compound when estimating $g$, which means that the dimensionality of $\vX$ is again a limiting factor.

%%%%%%%%%%%%%%%%%%%%%%%%%%%%%%%%%%%%%%%%%%%%%%
%\paragraph{Unsupervised learning as Supervised learning and Association Rules.} \\

%%%%%%%%%%%%%%%%%%%%%%%%%%%%
\paragraph{\cde reformulation.} There are other ways to reformulate the \cde, for example \cite{sugiyama_conditional_2010} propose a reformulation into a Least-Squares Density Ratio (\lsc) and \cite{meinshausen_quantile_nodate} into a quantile regression task.  Still others take advantage of a reformulation of the \cde  into a supervised learning regression task. Meanwhile, \rfcde (\cite{pospisil_rfcde_2018}) and \nnk (\cite{pospisil_nnkcde_2020}) are techniques that adapt methods that have proven to be effective to the \cde task.
Other methods, such as \dcde (\cite{disanto_photometric_2018}) and \flex (\cite{izbicki_converting_2017}), go further and design a surrogate regression task that can be run by  an off-the-shelf supervised learning method taking advantage of many mature and well-supported open-source projects, $e.g.$ scikit-learn (\cite{pedregosa_scikit-learn_2011}), \packpytorch (\cite{paszke_pytorch_2019}), \textit{fastai} (\cite{howard_fastai_2020}) , \textit{keras} (\cite{chollet_keras_2022}), \packtensorflow (\cite{martin_abadi_tensorflow_2015}), \CAT (\cite{prokhorenkova_catboost_2018}), \XGB (\cite{chen_xgboost_2016}), etc.  Our method implementation also has this advantage, since it can take as arguments any class that follows the scikit-learn $init/fit/predict\_proba$ API, which includes our off-the-shelf \MLP implementations, linear grid units (\GLU,\cite{gorishniy_revisiting_2021}), \ResBlock (\cite{gorishniy_revisiting_2021}) and self-normalizing networks (\SNN,\cite{klambauer_self-normalizing_2017}).\\

%%%%%%%%%%%%%%%%%%%%%%%%%%%%%%%
\paragraph{Neural networks and variational inference.}

Long before the current resurgence of interest in deep learning, there were already neural networks designed specifically to handle \cde, $e.g.$ 
\cite{bishop_mixture_1994} adressed the problem of estimating a probability density using Mixture Density Networks (\mdn).  More recently, Kernel Mixing Networks (\kmn, \cite{ambrogioni_kernel_2017}) are networks trained to estimate a family of kernels to perform the \cde task. Another famous variational inference technique is Flow Normalization (\nf, \cite{rezende_variational_2016}), designed to solve the problem of finding the appropriate approximation of the posterior distribution. A common challenge with most of these methods is scalability in terms of computation resources, which our experimental benchmark confirmed. A \packpython implementation of \mdn, \kmn and \nf is provided by the $freelunchtheorem$ package (\cite{noauthor_conditional_nodate}), which we included in our benchmark.

%%%%%%%%%%%%%%%%%%%%%%%%%%%%%%%%%%%%%%%  
\subsection{Our contributions}
%%%%%%%%%%%%%%%%%%%%%%%%%%%%%%%%%%%%%%%%%%%

Our method, called contrastive marginal discrimination, \ourmeth, combines several characteristics of the above methods but without their respective limitations. At a basic level, \ourmeth begins by reformulating the conditional density function into two factors, the marginal density function of $\vY$ and a ratio of density functions as follows
%$\pf{Y}$ and a ratio of density functions $\frac{\pf{X,Y}}{\pf{X}\pf{Y}}$ as follows
$$
\pf{Y\pCond X=x}(y) = \pf{Y}(y)\frac{\pf{X,Y}(x,y)}{\pf{X}(x)\pf{Y}(y)},\quad \forall (x,y) \in \pOmegX\times\pOmegY \text{ s.t. }\pf{X}(x)\neq0,\,\, \pf{Y}(y)\neq0
$$
We propose to estimate these two quantities separately. In most real applications of \cde, $Y$ is univariate or low-dimensional, while $X$ is not. In these cases, it is much easier to estimate $\pf{Y}$ rather than $\pf{X}$ and $\pf{X,Y}$. In this \ourpaper, we do not focus on how to choose the estimation method for  $\pf{Y}$ or  introduce new techniques to estimate $\pf{Y}$. By contrast, our experimental results show that out-of-shelf kernel estimators with default parameters perform very well on both simulated density models and real datasets, as expected for univariate distributions.\\

The core of our method is the reformulation of the ratio of the density functions $\pf{\vX,\vY}/\big(\pf{\vX}\pf{\vY}\big)$  into a contrast. In our method, the introduced noise in equation~\ref{eq:noisecontrastive} always corresponds to the density function $g(\cdot) = \pf{\vX}\pf{\vY}$. This is akin to the reformulation proposed by \cite{tibshirani_elements_nodate}, except that we only break the relationship between the two elements of the pair $(\vX,\vY)$ but not between each component of $\vX$ and $\vY$: $\pf{\vX,\vY}(\cdot) = \pf{\vX}(\cdot)\pf{\vY}(\cdot) \frac{\contrast(\cdot)}{1 - \contrast(\cdot)}$. To estimate the joint density $\pf{\vX,\vY}$ it would be necessary to estimate both $\pf{\vX}$ and $\pf{\vY}$. But when we apply Bayes' formula (\ref{eq:Bayes0}), we divide by $\pf{\vX}$, which disappears from the expression of $\pf{Y\pCond X}$, meaning we only need to estimate $\pf{\vY}$ and $\contrast$.\\

Like noise-contrastive methods, \ourmeth can leverage \textit{state-of-the-art} supervised learning techniques to perform \cde, especially neural networks.
Our numerical experiments reveal \ourmeth performances are far superior when using neural networks compared to other popular classifiers like \CAT, \XGB or Random Forest. Our benchmark also reveals that our method significantly outperforms in practice \rfcde, \nnk, \mdn, \kmn, \nf, \dcde, \flex and \lsc on most the density models and regression datasets included in our benchmark. Moreover, the \ourmeth reformulation enables us to train the binary classifier on a contrast training set much larger than the original dataset. Evermore, \ourmeth can easily take advantage of additional data. Unlabeled observations can be directly used to increase the size of the training set, without any drawbacks. Similarly, in the case where each observation is associated with more than one target value, they can all be included in the training dataset.\\

% Finally, we also exhibit in our experimental section how \ourmeth can also have applications in self-supervised learning even in contexts where the density is intractable or unproper, as should be expected since it is a contrastive method. In those cases, we can still leverage the ability of the discriminator to evaluate how related are a target value and an observation.\\
Our main contributions are as follows:
\begin{itemize}
    \item We introduce a reformulation of the \cde problem into a contrastive learning task which combines a binary classification task and a marginal density estimation task, which are both much easier than \cde. 
    \item We prove that given a training set of size $\SampSizexy$ it is always possible to generate a $\pIID$ training set of size $\ArrB{\frac{\SampSizexy }{2}}$ corresponding to the contrast learning task. We also prove that it is always possible to generate a training set of size at most $\SampSizexy^2$ in the non $\pIID$ case.
    \item We provide the corresponding construction procedures and the \packpython implementation. We also provide construction procedures to leverage additional marginal data and multiple targets per observations, which can improve performances significantly.
    \item We produce a benchmark of $9$ density models and $12$ datasets. We combine our method with a large set of classifiers and neural networks architectures, and compare ourselves against a large set of \cde methods. Our benchmark reveals that \mcd \textbf{outperforms all the existing methods on the majority of density models and datasets, sometimes by a very significant margin}.
    \item We provide a \packpython implementation of our method compatible with any \packpytorch module or \textit{scikit-learn} classifier, and the complete code to replicate our experiments.
\end{itemize}
\noindent\\

The rest of this \ourpaper is organised as follows: section~\ref{Sec:theory} provides a theoretical background for the reformulation. Section~\ref{Sec:experiments} provides the implementation details and evaluates our method on density models and regression datasets, comparing results with the methods implemented in the \packpython frameworks \packCDEtools\cite{dalmasso_conditional_2020,pospisil_cdetools_2022} and \packfreelunchtheorem \cite{noauthor_conditional_2022,rothfuss_conditional_2019}. Section~\ref{Sec:ablation} provides an ablation study of our method. Section~\ref{Sec:proofs} provides the proofs for theoretical results and the algorithms to construct the training dataset.\\

%% file: Experiments.tex
%%%%%%%%%%%%%%%%%%%%%%%%%%%%%%%%%%%%%%%%%%%%%%%%
\section{Experiments}\label{Sec:experiments}
%%%%%%%%%%%%%%%%%%%%%%%%%%%%%%%%%%%%%%%%%%%%%%%%

In this section we detail the implementation of \ourmeth{} in section~\ref{sec:imple} and provide a benchmark to compare \ourmeth{} to other available methods. All our experiments are done in \packpython and the random seed is always set such that the results of our method are fully reproducible. \\

We compare our method \mcd{} with other well-known methods presented in section~\ref{sec:benchmarkMethod} on both density models including two new ones, described in section~\ref{sec:expdens} and real dataset. Our results are displayed in sections~\ref{sec:resultsExperiments} and~\ref{sec:realExperiments}.

%Remark that \mcd can be used to perform to task: non-parametric estimation for any $x \in \pOmegX $ of the function $\pf{\vY \pCond \vX = x}$ and parametric estimation for any $(x,y) \in \pOmegX \times \pOmegY$ of the point-wise evaluation $\pf{\vY \pCond \vX = x}(y)$.

%We provide the method implementation to do both in section~\ref{sec:imple}.
%In section~\ref{sec:expdens}, we introduce known density models $\pf{\vY \pCond \vX = x}(\cdot)$. In section~\ref{sec:resultsExperiments}, we compare \mcd{} with other methods (see  on a grid of target values $\mathcal G_Y\subset \pOmegY$\\

%In our benchmark experiments, we introduce density models for which $\pf{\vY \pCond \vX = x}(\cdot)$ is known, but we compare \mcd with parametric estimation methods in a point wise fashion: for any observation $x \in \pOmegX$ of the test set, we evaluate $\pf{\vY \pCond \vX = x}$ and $\pfhat{\vY \pCond \vX = x}$ on a grid of target values in $\pOmegY$. The benchmark methodology is detailed in section ~\ref{sec:expdens}.

%%%%%%%%%%%%%%%%%%%%%%%%%%%%%%%%%%%%%%%%%%%%%%%%
\subsection{Method implementation}\label{sec:imple}
%%%%%%%%%%%%%%%%%%%%%%%%%%%%%%%%%%%%%%%%%%%%%%%%

\paragraph{Training.} First describe the procedure used for both parametric and nonparametric estimation, to train our estimator \mcd on  dataset corresponding  respectively to Framework ~\ref{DATA1}, ~\ref{DATA2} or ~\ref{DATA3}.  Table~\ref{tab:lookup} details which Construction to use in each Framework.\\

\noindent
\textbf{Training} \quad \begin{tabular}[t]{|l}
\begin{minipage}[t]{10cm}
    $(\text{Step}_1).$ Set $\ratio$.\\
    $(\text{Step}_2).$ Estimate $\pf{\vY}$ using $\{\vY_i\}_{i=1,\cdots,\SampSizexy}$ from $\mDxy$.\\
    $(\text{Step}_3).$ Generate $\mDwz{\SampSizewz}$. \\
    $(\text{Step}_4).$ Train either a regressor or a binary classifier on $\mDwz{\SampSizewz}$.
\end{minipage}
\end{tabular}
\\
The estimators obtained in steps 2 and 4 are called respectively the \textbf{marginal estimator} $\pfhat{\vY}$ and the \textbf{discriminator} $\contrasttild$. Note that for any $(x,y) \in \pOmegX \times \pOmegY$, we have $\,\contrasttild(x,y)\in [0,1]$ while $\contrast(x,y) \in [0,1)$. To obtain an appropriate prediction, we introduce a thresholding constant $\epsilon = 10^{-6}$ and set $$\contrasthat(x,y) = \min(\contrasttild(x,y), 1 - \epsilon) \in [0,1 - \epsilon] \subset [0,1).$$
\begin{table}[http]\label{tab:lookup}
\centering
\begin{tabular}{|l|l|c|c|}
\hline
Framework & Available datasets & \pIID\,\,samples & \pUID\,\,samples \\
\hline
\hline
Framework~\ref{DATA1} & $\mDxy$ & Construction~\ref{const:DwzNiid} & Construction~\ref{const:DwzNid} \\
\hline
Framework~\ref{DATA2}& $\mDxy, \mDx, \mDy$ & Construction~\ref{const:DwzNiidAD} & Construction~\ref{const:DwzNidAD} \\
\hline
Framework~\ref{DATA3} & $\mDxyM$ & Not applicable & Construction~\ref{const:DwzNidFram3} \\
\hline
\end{tabular}
\caption{Look-up table to determine the appropriate construction corresponding to each framework.}
\end{table}

Next, underline that \ourmeth{} can be used to perform two different tasks:
\begin{itemize}
    \item Nonparametric estimation of the conditional density
    $\pf{\vY \pCond \vX = x}(\cdot)$ for all $x \in \pOmegX $.
    \item Pointwise estimation of the conditional density
    $\pf{\vY \pCond \vX = x}(y)$ for any $(x,y) \in \pOmegX \times \pOmegY$.
\end{itemize}
Indeed, using Fact~\ref{fac:MCF} we have: $\forall (x,y) \in \pOmegX \times \pOmegY$, \begin{equation}\label{eq:plugin}
\pfhat{\vY \pCond \vX = x}(y) = \pfhat{\vY}(y) \frac{\contrasthat(x,y)}{1 - \contrasthat(x,y)}\frac{1 - \ratio}{\ratio}.
\end{equation}
This implies that it is sufficient to have estimators of both $\pf{\vY}$ and $\contrast$ to have a point estimate of $\pf{\vY\pCond \vX = x}(y)$. We may also deduce the literal expression of $\pfhat{\vY \pCond \vX = x}(\cdot)$, the nonparametric estimate of the conditional density, provided we know the literal expressions of  $\,\widehat{\contrast}(x,\cdot)$ and $\pfhat{\vY }(\cdot)$. It is the case for $\widehat{\contrast}${} in Deep Learning, as we can write the literal expression of $\,\widehat{\contrast}(x,\cdot)$ from the \NeurN parameters learned in the learning step and a value $x$. %(see chapter 2).  
Under certain assumptions on $\widehat{\contrast}$,\, $\pfhat{\vY \pCond \vX = x}(\cdot)$ is a true density (\cite{gutmann_noise-contrastive_2010}). \\

\paragraph{Prediction.} For parametric pointwise estimation, at test time, given any new observation $x \in \pOmegX$, for any chosen target value $y \in \pOmegY$, we can estimate $\pf{\vY \pCond \vX = x}(y)$  the value of the probability density function evaluated on $(x,y)$:\\\\
\noindent
\textbf{Prediction} \quad \begin{tabular}[t]{|l}
\begin{minipage}[t]{10cm}
$(\text{Step}_1).$ Evaluate $\contrasttild(x,y)$. \\
$(\text{Step}_2).$ Apply thresholding: $\contrasthat(x,y) = \min(\contrasttild(x,y), 1 - \epsilon)$.\\%Note that in practice this should almost always be the case when using a neural network classifier with the sigmoid output activation function. \\
$(\text{Step}_3).$ Evaluate $\pfhat{\vY}(y)$.\\
$(\text{Step}_4).$Plug in $\pf{\vY \pCond \vX = x}(y)$ by applying equation~\ref{eq:plugin} given above.
\end{minipage}
\end{tabular}\\

\noindent Remark that if the discriminator is a classifier, the predicted value should be the probability of class 1, $i.e.$ $\pP[\vZ = 1\pCond \vW = (x,y)]$.

%%%%%%%%%%%%%%%%%%%%%%%%%%%%%%%%%%%%%%%%%%%%%%%%
\paragraph{Choice of parameters.} There are 4 major choices to make when implementing our method: the marginal density  estimator method, the discriminator method, the construction and the contrast ratio $\ratio$.\\

\noindent$\bullet$ \textbf{Marginal density estimator.} Since the point of our method is to provide an estimation of the conditional density in cases where the marginal density is relatively easy to estimate (meaning $\pOmegY \subseteq \eR$), we pick a simple yet effective technique, the univariate kernel density estimation \textit{KDEUnivariate} provided in the $statsmodels$ package. We always keep the default parameters, $i.e.$ gaussian kernels, bandwidth set using the normal reference, and the fast Fourier transform algorithm to fit the kernels.\\
%
%Regarding Neural Networks, we implemented $5$ different architectures: MultiLayerPerceptron(\MLP), ResNet (\ResBlock) \cite{gorishniy_revisiting_2021}, Gated Linear Unit(\GLU)\cite{gorishniy_revisiting_2021}, Self Normalizing Networks (\SNN)\cite{klambauer_self-normalizing_2017} and an \MLP without Drop-Out nor Batch-Norm (\MLP:\NoDO) \cite{srivastava_dropout_2014}. For the most part, we used the same hyper-parameter values as in chapter 4, as these appeared appropriate in our early investigations.\\

%We also evaluated our method with a set of popular classifiers implemented in \packpython, including Random Forests (\RF) and elastic-net (\textit{scikit-learn} \cite{pedregosa_scikit-learn_2011}), Gradient Boosting Decision Trees (\XGB \cite{chen_xgboost_2016}, \CAT \cite{prokhorenkova_catboost_2018}, \LGBM \cite{ke_lightgbm_2017}), and Feed-Forward Neural Networks (\packpytorch  \cite{paszke_pytorch_2019}).\\
\begin{table}[http!]\label{tab:methods}
\centering
\footnotesize
\begin{tabular}{|l|l|}
\hline
Neural Networks architectures & Other supervised learning classifiers \\
\hline
\hline
MultiLayerPerceptron (\MLP) & Random Forests (\RF) \\
\hline
\MLP w/o Drop-Out nor Batch-Norm (\MLP:\NoDO) & Elastic-net \\
\hline
ResNet (\ResBlock) \cite{gorishniy_revisiting_2021} & \XGB  \cite{chen_xgboost_2016} \\
\hline
Gated Linear Unit (\GLU) \cite{gorishniy_revisiting_2021} & \CAT \cite{prokhorenkova_catboost_2018} \\
\hline
Self Normalizing Networks  (\SNN) &  \LGBM \cite{ke_lightgbm_2017}) \\
\cite{klambauer_self-normalizing_2017} & \\
\hline
\end{tabular}
\normalsize
\caption{List of discriminator methods evaluated with \mcd.}
\label{table:classifiers}
\end{table}
\noindent$\bullet$ \textbf{Marginal contrast discriminator.}
We evaluate the performance of $\mcd$ combined with Neural Networks, Decision Tree based classifiers and Logistic Elastic-net (see table~\ref{tab:methods} for the exhaustive list). %Deep Learning architectures used are described in Chapter 4.\\ %\textcolor{magenta}{ref TODO chapitre ref} \\
%We restrained from assessing the impact of data augmentation techniques and other supervised learning schemes since the aim of this chapter is not to see how each supervised learning scheme can impact our method but rather to assess its performances when compared with common methods. This could be an interesting avenue of research for future works.\\

\noindent$\bullet$  \textbf{Dataset construction and ratio:}
We compare in our ablation study (Section~\ref{Sec:ablation}) the Constructions~\ref{const:DwzNiid}, \ref{const:DwzNid}, \ref{const:DwzNiidAD}, \ref{const:DwzNidAD} and  \ref{const:DwzNidFram3} in Frameworks~\ref{DATA1}, ~\ref{DATA2} and ~\ref{DATA3}. Following the findings of the ablation study, we use Construction~\ref{const:DwzNid} and set $\ratio = 0.05$ in other experiments.

%%%%%%%%%%%%%%%%%%%%%%%%%%%%%%%%%%%%%%%%%%%%%%%%
\subsection{Other benchmarked methods and Application Programming Interface (API)}
\label{sec:benchmarkMethod}
%%%%%%%%%%%%%%%%%%%%%%%%%%%%%%%%%%%%%%%%%%%%%%%%

There is a small number of \cde{} methods for which a readily available open source implementation in \packpython exists. One notable difficulty when introducing a new \cde{} package is that there is no gold standard API in \packpython on top of which new packages can build upon. Most existing implementations are standalone, with their own unique syntax for common functions. We choose to include in our benchmark the methods provided by two of the most mature \packpython projects, the \FLT \packgithub repository by \cite{rothfuss_conditional_2019, noauthor_conditional_2022}, and a network of packages created by \cite{dalmasso_conditional_2020} and \cite{pospisil_cdetools_2022}.\\

\noindent The \FLT \packgithub provides an implementation of \kmn, \nf, \mdn and \lsc. The\FLT \packgithub  has a quite consistent API across all provided methods. Notably, \cite{rothfuss_conditional_2019, noauthor_conditional_2022}, also provide implementations for several statistical models (\dEcon, \dArma, \dJump, \dLinG, \dStud, \dSkew and \dGMixt), which we include in our benchmark (see the density model section~\ref{sec:expdens}).\\

\noindent Meanwhile, the project of \cite{dalmasso_conditional_2020} and \cite{pospisil_cdetools_2022} is built around the \packCDEtools \packgithub repository and consists of several repository of varying maturity and ease of use corresponding to each method they implemented (\rfcde, f-$\rfcde$, $\flex$, \dcde and \nnk).  Notably, they also provide implementations in Java and R for some of these methods. Their implementation of \dcde allows custom $pytorch$ and $tensorflow$ architectures to be plugged-in, which gave us the opportunity to adapt the architectures and training schemes used with \mcd to \dcde. As such, the comparison between \mcd and \dcde is done on equal ground.\\

\noindent In total, we include in our benchmark $10$ other \cde methods: \nnk, \nf, \lsc, \mdn, \kmn, \dcde, \rfcde, \textbf{f}-\rfcde, \flex:\NeurN and \flex:\XGB.\\

\noindent Although it is not the main goal of this work, we provide an overhead over these two projects and our method to facilitate the comparison between them. Each evaluated method is encapsulated in a class which inherits the same unique API from the parent class \textit{ConditionalEstimator}, with same input and output format and global behavior. Similar to \textit{scikit-learn}, the estimator is an instance of the class, with hyper-parameters provided during initialisation ($\_\_init\_\_)$, and the observation matrix and target vector provided (as $numpy$ arrays) when calling the $fit$ function. At predict time however, the estimator prediction is a function called \textit{pdf\_from\_X} which predicts the probability density function on a grid of target values. This package overhead allows us to compare all methods on equal ground: For density models all methods are trained on the same sampled training set. We also use the same grid of target values and test set of observations to evaluate the probability density function of all compared methods. Likewise, for real-world datasets, we use the same dataset train-test splits for all compared methods.\\ 

%%%%%%%%%%%%%%%%%%%%%%%%%%%%%%%%%%%%%%%%%%%%%%%%
\subsection{Estimation of theoretical models}
\label{sec:expdens}
%%%%%%%%%%%%%%%%%%%%%%%%%%%%%%%%%%%%%%%%%%%%%%%%

We first evaluate our method on the core task it aims to handle on theoretical models: numerically estimating  a conditional density function with respect to a new observation. We choose to evaluate the quality of the prediction empirically: for each predicted and target functions, we evaluate for a grid of target values the \textbf{empirical Kullback-Leibler divergence} (\pKL).%, Jensen-Shannon(\pJS), \pHL and \pLt norm: %Let us denote , and let $y_{min}$ and $y_{max}$ be the estimated bounds of the support of $\vY$ $\{y_i\}_{i = 1, \cdots, 9999} = \{y_{min} + \frac{i * (y_{max}-y_{min})}{10000}\}_{i = 1, \cdots, 9999}$ the grid of target values for which the probability density function will be evaluated. 

%%%%%%%%%%%%KL KULLBACK divergence definition

% , $\pJS$ distance, $\pHL$ distance and $\pLt$ norm as follows:
%  \begin{align*}
%      \pKL(f,g)&:= f \times ln\left( \frac{f}{g}\right). \\
%      \pJS(f,g)&:= \frac{1}{2}\pKL(f,g) + \frac{1}{2}\pKL(g,f) \\
%      \pHL(f,g)&:= (\sqrt{f}-\sqrt{g})^2 \\
%      \pLt(f,g)&:= (f-g)^2.
%  \end{align*}
\paragraph{Density models} The \FLT package provides $7$ conditional densities implementations for which we have a function to generate a training dataset and a function to evaluate the theoretical density function on a grid of target values given an observation. These 7 models are \dEcon, \dArma, \dJump, \dLinG, \dStud, \dSkew and \dGMixt. We refer to the \FLT \packgithub documentation \cite{noauthor_conditional_2022} for a detailed description of each model. Although these conditional density models cover a diverse set of cases, we do however introduce two other density models to illustrate the specific drawbacks of some benchmarked methods.\\

 \begin{mdframed}
\begin{mymodel}\label{Basic}[\dBasic]:
Let $\DimInput = 10$ and fix $\DimInput$ coefficients $\beta = \{ \beta_j \}_{j = 1, \cdots, \DimInput}$ drawn independently at random, such that $\forall j = 1, \cdots, \DimInput,$ $\beta_j$ is uniformly distributed over $(0,1)$, $i.e.$ $\beta_j \sim \pUU(0,1)$.\\

Construct now our first density model
\begin{itemize}
    \item Let $\vX \sim \pNN(\Zeros_\DimInput, \Id_\DimInput)$ be a gaussian vector. 
    \item Let $\vY \in \eR$ be a random variable such that $\vY = \vX^\transp \beta + \sigma \epsilon$ where $\epsilon \sim \pNN(0,1)$ and $\epsilon$ and $\vX$ independent. 
\end{itemize}
\end{mymodel}
\end{mdframed}
$\dBasic$ is a very simple linear model included to check that sophisticated methods which can estimate complex models are not outperformed in simple cases. We also add a second model, $\dAsym$, which corresponds to $\dBasic$ with a simple modification: we use asymmetric noise ($|\epsilon|$ instead of $\epsilon$).\\
 \begin{mdframed}
\begin{mymodel}\label{Asym}[\dAsym]:
Let $\beta$, $\vX$ and $\epsilon$ be as in $\dBasic$\ref{Basic}.
In our second density model, $\vY$ is as follows:
$$
\vY = \vX^{\transp} \beta + \sigma |\epsilon|.
$$
Here, $|\cdot|$ denotes the absolute value. 
\end{mymodel}
//
\end{mdframed}
 The major difficulty is that the support of the conditional density of $\vY$ with respect to $\vX$ differs from the support of the marginal density of $\vY$ (which is $\pOmegY = \eR$) and depends on the observation $\vX$.

%%%%%%%%%%%%%%%%%%%%%%%%%%%%%%%%%%%%%%%%%%%%%%%%
\subsection{Results on density models}
\label{sec:resultsExperiments}
%%%%%%%%%%%%%%%%%%%%%%%%%%%%%%%%%%%%%%%%%%%%%%%%
\paragraph{Evaluation protocol.} 
We use $9$ density models (section~\ref{sec:expdens}) as ground truth, on which we evaluate the \mcd{} with various discriminators and the methods presented in section~\ref{sec:benchmarkMethod}.% Kernel Mixture Networks (\kmn, Normalizing flows (\nf), Mixture density networks (\mdn), Least-square conditional (\lsc), \rfcde, f-$\rfcde$, $\flex$, \dcde and Nearest Neighbor Kernels (\nnk). 
To generate $\mDxy$ for each density model, we sample $\SampSizexy = 100$ observations and for each observation we sample one target value using the conditional law of the density model. We train all benchmarked methods on this same dataset. Next, we sample $n_{test} = 100$ observations from the density model to generate a test set. We also generate a unique grid of $10000$ target values, spread uniformly on $\pOmegY$. For each observation of the test set, we evaluate the true conditional density function on the grid of target values. Then, for all benchmarked method, we estimate the conditional density for each observation on that same grid of target values and evaluate the empirical Kullback-Leibler (\pKL)  divergence defined below.\\

\begin{mdframed}
\textbf{Empirical Kullback-Leibler divergence}.
\noindent\\ Set $\delta= 10^{-6}$ a numerical stability constant. Let $x \in \eR^{\DimInput}$ be an observation of the test set $\mathcal{D}_{\textit{test}}$  and $y\in \eR$ be a point on $\mathcal{G}$, a grid of target values to be estimated. Then, for the evaluation of the target value $f_{x}(y)= \max(\pf{\vY \pCond \vX = x}(y),\delta)$  and the predicted value $g_{x}(y) = \max(\pfhat{\vY \pCond \vX = {x}}(y),\delta)$,  we define the empirical $\pKL_{\! \delta }$ divergence as follows: 
\begin{equation}
\pKL := \pKL_{\! \delta }(f \,\|\, g)= \sum_{x\, \in\, \mathcal{D}_{\textit{test}}} \sum_{y\, \in \, \mathcal{G}} f_{x}(y) \times \ln\left( \frac{f_{x}(y)}{g_{x}(y)}\right).
\end{equation}
\end{mdframed}

%%%%%%%%%%%%%%%%%%%%%%%%
\paragraph{Benchmark results.} 
\noindent\\
\noindent \ding{226} We first combine the \mcd method with a classic Multi Layer Perceptron (\MLP) as discriminator and a kernel estimator as marginal density estimator, named \mcd:\MLP. Table~\ref{table:density} depicts the global performance in terms of empirical $\pKL$ divergence 
over 9 models of \mcd:\MLP compared to 10 others methods, described in section~\ref{sec:benchmarkMethod}.
%
%\textcolor{magenta}{of the \mcd method combined with a classic Multi Layer Perceptron (\MLP) as discriminator and a kernel estimator as marginal density estimator compared to 10 others methods described in section~\ref{sec:benchmarkMethod}}. 

\begin{itemize}
    \item The main take away is that in $6$ out of $9$ cases, \mcd:\MLP outperforms all others. 
    \item On $3$ density models, \dBasic, \dLinG and \dStud, the \pKL empirical divergence is less than half of the second best method. Meanwhile, on \dEcon, \nf and \mcd:\MLP share the first place.
    \item When \mcd:\MLP is outperformed, which corresponds to \dJump and \dSkew, the best performing methods are \nf and \mdn. Otherwise, the second best performing method is either \nf or \nnk.
\end{itemize}

\begin{table}[http]
\scriptsize
\begin{tabular}{l c c c c c c c c c c c}
Empirical \pKL & \rot{\mcd:\MLP} & \rot{\nnk} & \rot{\nf} & \rot{\lsc}& \rot{\mdn} & \rot{\kmn} & \rot{\dcde} & \rot{\rfcde} & \rot{f-\rfcde} & \rot{\flex:\NeurN}&\rot{\flex:\XGB}  \\
\hline
\vspace{-0.25cm}
 & & & & & & & & & & & \\
\dArma & 0.573 & 0.929 & \textbf{0.196} & 0.408 & \underline{0.29} & 0.312 & 1.754 & 0.529 & 0.526 & 1.201 & 2.226 \\
\dAsym & \textbf{0.158} & \underline{0.245} & 0.498 & 0.882 & 0.437 & 0.338 & 0.666 & 0.324 & 0.328 & 0.359 & 0.483 \\
\dBasic & \textbf{0.009} & \underline{0.087} & 0.313 & 0.473 & 0.195 & 0.167 & 0.317 & 0.139 & 0.139 & 0.174 & 0.116 \\
\dEcon & \textbf{0.006} & 0.01 & \textbf{0.006} & 0.021 & \underline{0.013} & 0.022 & 0.068 & 0.049 & 0.045 & 0.034 & 0.048 \\
\dGMixt & \textbf{0.005} & \underline{0.008} & 0.012 & 0.023 & 0.016 & 0.018 & 0.127 & 0.026 & 0.028 & 0.048 & 0.023 \\
\dJump & \textbf{1.352} & \underline{1.632} & 4.481 & 9.371 & 4.576 & 5.568 & 22.45 & 10.45 & 10.45 & 6.347 & 4.363 \\
\dLinG & \textbf{0.189} & 1.742 & \underline{0.868} & 3.15 & 2.318 & 2.892 & 14.71 & 15.88 & 15.88 & 13.75 & 3.571 \\
\dStud & \textbf{0.141} & \underline{0.301} & 6.238 & 9.09 & 3.109 & 3.136 & 7.583 & 2.363 & 2.363 & 1.686 & 0.821 \\
\dSkew & 0.722 & 0.089 & \underline{0.019} & 0.1 & \textbf{0.014} & 0.036 & 18.94 & 0.551 & 0.551 & 0.636 & 2.255 \\
\hline
\end{tabular}
\caption{Evaluation of the empirical \pKL divergence of different \cde methods, for $9$ density models with $\SampSizexy=100$. \textbf{Best performance} is in bold print, \underline{second best performance} is underlined. Lower values are better. Column $2$ corresponds to the performance of $\mcd$ combined with the classic \MLP. Columns $3$ to $12$ show the results of the $10$ other benchmarked methods.}
\label{table:density}
\end{table}
\noindent \ding{226} We also assess the performance of \mcd combined with other popular supervised learning methods. Table~\ref{table:discriminator} depicts the performance of \mcd with various discriminators on the same benchmark of density models. The classifiers included are listed in Table~\ref{table:classifiers} and described in section~\ref{sec:imple}.

\begin{itemize}
    \item In density model \dArma where \mcd:\MLP is outperformed by other methods, simply removing the Batch-Normalization and Drop-Out is sufficient to obtain the best performance.
    \item In density models \dBasic, \dEcon, \dGMixt, \dJump, \dLinG and \dStud, the results for \mcd:\MLP are very close to other \NeurN architectures performances. This means that in most cases, \mcd does not require heavy tuning to perform well.
    \item The best discriminator besides \NeurN is always either \CAT or \Enet.
    \item \NeurN discriminators are outperformed by other classifiers in only one case, the \dSkew density model. This corresponds to the density model included in our benchmark where all versions of \mcd are outperformed by another method.

\end{itemize}

\begin{table}[http]
\scriptsize
\begin{tabular}{l c c c c c c c c c c c}
Empirical \pKL & \rot{\MLP} & \rot{\SNN} & \rot{\GLU} & \rot{\ResBlock} & \rot{\NoDO} & \rot{\CAT} & \rot{\Enet} & \rot{\XGB} & \rot{\LGBM} & \rot{\RF} & \rot{\XRF} \\
\hline
\vspace{-0.25cm}
 & & & & & & & & & & & \\
\dArma & 0.573 & 1.047 & 0.731 & \underline{0.224} & \textbf{0.1} & 0.241 & 1.09 & 0.598 & 0.953 & 2.265 & 4.469 \\
\dAsym & 0.158 & \textbf{0.05} & 0\textbf{0.05} & \underline{0.055} & 0.318 & 0.2 & 0.262 & 0.274 & 0.28 & 0.279 & 0.319 \\
\dBasic & \underline{0.009} & \textbf{0.008} & 0.01 & \underline{0.009} & 0.108 & 0.059 & 0.083 & 0.113 & 0.098 & 0.094 & 0.106 \\
\dEcon & \underline{0.006} & 0.007 & \textbf{0.005} & \textbf{0.005} & \underline{0.006} & 0.012 & 0.016 & 0.06 & 0.07 & 0.305 & 0.444 \\
\dGMixt & \textbf{0.005} & \underline{0.006} & \underline{0.006} & \textbf{0.005} & 0.007 & 0.01 & 0.007 & 0.057 & 0.061 & 0.297 & 0.422 \\
\dJump & \textbf{1.352} & \underline{1.353} & \textbf{1.352} & \textbf{1.352} & \textbf{1.352} & 1.485 & \textbf{1.352} & 5.694 & 5.11 & 10.17 & 19.99 \\
\dLinG & \textbf{0.189} & \textbf{0.189} & \underline{0.19} & \underline{0.19} & \underline{0.19} & 1.274 & \underline{0.19} & 9.33 & 10.99 & 67.46 & 94.76 \\
\dStud & \textbf{0.141} & \textbf{0.141} & \textbf{0.141} & \textbf{0.141} & \textbf{0.141} & \underline{0.236} & \textbf{0.141} & 1.58 & 1.027 & 0.912 & 1.617 \\
\dSkew & 0.722 & 0.796 & 0.669 & 0.356 & 0.155 & \textbf{0.083} & 0.833 & 0.223 & \underline{0.12} & 5.036 & 7.289 \\
\hline
\end{tabular}
\caption{Evaluation of the empirical \pKL divergence of different \mcd{} discriminators, for $9$ density models with $\SampSizexy=100$. \textbf{Best performance} is in bold print, \underline{second best performance} is underlined. Lower values are better. Columns $2$ to $6$ correspond to the performance of $\mcd$ combined with \NeurN architectures. Columns $7$ to $12$ show the results of $\mcd$ combined with other popular classifiers.}
\label{table:discriminator}
\end{table}

\paragraph{Impact of dimensionality.} 
\noindent\\
\noindent \ding{226} Then, we check if in cases where \mcd outperforms all others, our method maintains its good performances when $\DimInput$, the number of features, changes. Table~\ref{table:features} depicts the impact of the dimensionality of $\vX$ on the performances of each method in \dBasic, the density model where the performance gap in our benchmark between \mcd and other methods is the largest.

\begin{itemize}
    \item Performances decrease across the board when $\DimInput$ increases. Although \dBasic is a setting where a larger $\DimInput$ corresponds to a higher signal to noise ratio, this is not enough to counter the curse of dimensionality.
    \item \mcd:\MLP outperforms all others at all ranges of $\DimInput$. Note that at $\DimInput = 300$, the empirical \pKL of \mcd:\MLP is equivalent to that of $\pf{\vY}$, the marginal density of the target value. This means that in this high-dimensional case, where \mcd:\MLP is not able to capture the link between $\vX$ and $\vY$, it does not overfit the training set, and instead takes a conservative approach.
    
    \item \nnk maintains good results across the board. Meanwhile, \mcd combined with \CAT performs almost as well as \mcd:\MLP when $\DimInput = 3$, but its relative performances are average at best when $\DimInput = 300$. 
\end{itemize}

\begin{table}[http]
\centering
\scriptsize
\begin{tabular}{|c||c|c|c|c|c|c|c|}
\hline
\vspace{-0.24cm}
 & & & & & & & \\
\multirow{2}{*}{Empirical \pKL}& \mcd & \mcd & \multirow{2}{*}{\nnk} & \flex & \multirow{2}{*}{\kmn} & \multirow{2}{*}{\rfcde} & \flex \\
& \MLP & \CAT & & \XGB & & & \NeurN \\
\hline
\hline
\vspace{-0.24cm}
 & & & & & & & \\
$\DimInput = \,\,\,\,3$ & \textbf{0.008} & \underline{0.009} & *0.022* & 0.093 & 0.067 & 0.037 & 0.094 \\
$\DimInput = \,\,10$ & \textbf{0.036} & \underline{0.042} & *0.063* & 0.076 & 0.096 & 0.105 & 0.106 \\
$\DimInput = \,\,30$ & \textbf{0.115} & 0.202 & \underline{0.154} & 0.196 & *0.181* & 0.238 & 0.22 \\
$\DimInput = 100$ & \textbf{0.162} & *0.224* & \underline{0.173} & 0.264 & 0.401 & 0.238 & 0.234 \\
$\DimInput = 300$ & \textbf{0.244} & 0.308 & *0.282* & 0.302 & 0.304 & 0.507 & \underline{0.253} \\
\hline
\end{tabular}
\caption{ Evaluation of the \pKL divergence values for various feature sizes $\DimInput$, on the \dBasic density model, with $\SampSizexy= 100$. \textbf{Best} performance is in bold print, \underline{second best} performance is underlined, *third best* performance is between asterisks. Lower is better. Methods depicted achieve top 4 performances for at least one feature size regime.}
\label{table:features}
\end{table}

\paragraph{Execution time and scalability.} 
\noindent\\
\noindent \ding{226}
Next, we evaluate the scability with respect to the size of the dataset of \mcd combined with either \MLP or \CAT, two classifiers known to be efficient but slow. We include as reference other methods belonging to the same categories, meaning those based on supervised learning \NeurN and Decision Trees respectively. We also include \nnk and the methods based on variational inference, as they perform well in our benchmark. Table~\ref{table:timing} depicts the computation time of \mcd and other \cde methods for $\SampSizexy=30,100,300$ and $1000$, on \dBasic, the density model where it is most beneficial to use \mcd instead of another method. The reported computation times include all steps (initialization, training, prediction). For \mcd, this includes both the time taken by the discriminator and the time taken by the estimator. Note also that for \mcd, we use the ratio $\ratio= 0.05$, meaning the actual training set size is $20$ times larger.

\begin{itemize}
    \item The main take away is that \mcd mostly scales like its discriminator would in a supervised learning setting.
    \item Regarding methods based on supervised learning with neural networks, we compare \mcd:\MLP with \flex:\NeurN and \dcde. \flex:\NeurN is much faster than \dcde and \mcd:\MLP. When using equivalent architectures, \dcde is slower than \mcd when $\SampSizexy = 1000$ (which corresponds to $\SampSizewz = \frac{\SampSizexy}{\ratio} = 20000$ for \mcd). This can be partly explained by the fact that \dcde uses a transformation of the target which corresponds to an output layer width of $30$, while $\mcd$ increases the input size of the network by $1$, since observations and target values are concatenated.
    \item Regarding methods based on supervised learning with decision trees, we compare \mcd:\CAT with \rfcde and \flex:\XGB. \rfcde is the only method which does not leverage $GPU$ acceleration, yet, it is faster than \mcd:\CAT and \flex:\XGB.
    
    \item Among the best performing methods in our benchmark, \nnk is by far the fastest. Meanwhile, the $3$ variational inference methods included, \mdn, \kmn and \nf, are much slower, which is to be expected. 
\end{itemize}

\begin{table}[http]
\centering
\tiny%\vline~ 
\begin{tabular}{|c||c|c|c||c|c|c||c||c|c|c|}
\hline
Category&\multicolumn{3}{|c||}{Decision Tree Based Methods}&\multicolumn{3}{|c||}{\NeurN based Methods}& Kernel & \multicolumn{3}{|c|}{Variational inference}\\
\hline
\vspace{-0.22cm}
 & & & & & & & & &  \\
Method & \rfcde & \mcd & \flex & \flex & \mcd & \dcde & \nnk &  \mdn & \nf & \kmn \\
\hline
\vspace{-0.22cm}
 & & & & & & & & &  \\
Based on: & \RF & \CAT & \XGB & \NeurN & \MLP & \MLP & N.Neighbor & \NeurN & \NeurN & \NeurN \\
\hline
\hline
$\SampSizexy = \,\,\,\,30$ & 0.044 & 0.46 & 9.574 & 0.018 & 3.133 & 1.943 & 0.01 & 35.46 & 50.74 & 47.92 \\
$\SampSizexy = \,\,100$ & 0.149 & 0.369 & 9.57 & 0.017 & 3.149 & 2.11 & 0.022 & 35.45 & 50.72 & 102.7 \\
$\SampSizexy = \,\,300$ & 0.448 & 0.446 & 13.19 & 0.017 & 3.132 & 3.225 & 0.04 & 35.65 & 50.62 & 150.6 \\
$\SampSizexy = 1000$ & 0.952 & 0.689 & 25.24 & 0.02 & 3.296 & 6.388 & 0.043 & 35.73 & 51.12 & 143.6 \\
\hline
\end{tabular}
\caption{Training Time in seconds for various training set sizes $\SampSizexy$, on the \dBasic density model. Row $1$ corresponds to the method category, row $2$ corresponds to the benchmarked \cde method and row $3$ corresponds to the underlying supervised learning method used. Column $1$ corresponds to the training set size. Columns $2$, $3$ and $4$ correspond to methods which leverage a supervised learning method based on Decision trees. Columns $5$, $6$ and $7$ correspond to methods which leverage a supervised learning method based on \NeurN. Column $8$ corresponds to nearest neighbors kernels. Columns $9$, $10$ and $11$ correspond to three variational inference methods.}
\label{table:timing}
\end{table}

%%%%%%%%%%%%%%%%%%%%%%%%%%%%%%%%%%%%%%%%%%%%%%%%
\subsection{Real-world datasets}
\label{sec:realExperiments}
%%%%%%%%%%%%%%%%%%%%%%%%%%%%%%%%%%%%%%%%%%%%%%%%
%%%%%%%%%%%%%%%%%%%%%%%%%%%%%%%%%%%%%%%%%%%%%%%%
\paragraph{Dataset origins and methodology.}
\noindent\\
\noindent \ding{226} We include in our benchmark $12$ datasets, taken from two sources: 
\begin{itemize}
    \item The \packCDEtools{} framework provides the dataset "Teddy" (see \cite{beck_realistic_2017} for a description). We follow the pre-processing used in the given packages.
    \item The \FLT{} framework provides $2$ toy datasets, $7$ datasets from the UCI (\cite{asuncion_uci_2007} repository, and one dataset from the kaggle plateform \cite{noauthor_kaggle_nodate} which includes $2$ targets, for a total of $11$ datasets. Here again, we follow the pre-processing used in the given package.
\end{itemize}

%%%%%%%%%%%%%%%%%%%%%
\paragraph{Evaluation protocol.}
\noindent\\
\noindent \ding{226}
For all datasets, we standardize the observations and target values, then we perform the same train-test split for all methods benchmarked  (manually setting the random seed for reproductibility). Because datasets are of varying sizes and as some methods are extremely slow for large datasets, we only take a subset of observations to include in the training set, doing the split as such. Let $n_{\max}$ be the original size of the dataset, the trainset is of size $\SampSizexy = \min(300,\ArrB{\SampSizexy_{\max} \times 0.8})$ and the test set is of size $n_{test} = \min(300,n_{\max}-\SampSizexy)$.\\
\noindent  One challenge when benchmarking \cde methods is that real-world datasets almost never provide the conditional density function associated to an observation, but instead only one realisation. This means that the usual metrics for \cde (eg.: \pKL) cannot be evaluated on these datasets. We nonetheless evaluate our method on real-world datasets, using the negative log-likelihood metric, denoted \pNLL, to compare the estimated probability density function against the target value. 

\begin{mdframed}
\textbf{Empirical Negative Log-likelihood}.
\noindent\\
Set $\delta= 10^{-6}$ a numerical stability constant. Let $(x,y) \in \eR^{\DimInput}\times\eR $ be a sample from the test set $\mathcal{D}_{\textit{test}}$, and $g_{x}(y) = \max(\pfhat{\vY \pCond \vX = {x}}(y),\delta)$ be the predicted value, we define the \pNLL metric as follows:
\begin{equation}
\pNLL := \pNLL_{\! \delta }(g)= - \!\!\!\!\sum_{(x,y)\, \in\, \mathcal{D}_{\textit{test}}} \! \ln\left( g_{x}(y)\right).
\end{equation}
\end{mdframed}

%%%%%%%%%%%%%%%%%%%%%%%%%%%%%%%
\paragraph{Results.}
\noindent\\
\noindent \ding{226} Table~\ref{table:real} depicts the global performance in terms of negative log-likelihood for the $12$ datasets.
\begin{itemize}
    \item The main take away is that this time, \mcd:\MLP outperforms existing methods in $7$ out of $12$ cases, including the popular datasets BostonHousing and Concrete.
    \item On the WineRed and WineWhite datasets, \mcd lags far behind \nnk and \flex. One possible cause is that for these two datasets, the target variable $\vY$ takes discrete values: $\pOmegY = \{3,4,5,6,7,8\}$, which does not correspond to the regression task for which \mcd was designed.
    \item Besides \mcd, the top performing methods are \nnk, \nf, \flex (both versions) and \kmn.
    \item Regarding the choice of the discriminator, \mcd:\MLP is not outperforming \mcd:\CAT to the same extent it does on density models. The difference in negative log-likelihood between \mcd:\MLP and \mcd:\CAT is below $0.1$ in $7$ out of $12$ cases.
    \item Besides, \mcd:\CAT{} obtains Top~ 2 performances in $3$ of the $5$ cases where \mcd:\MLP{} is outperformed by other methods. Notably, on the Yacht dataset where \mcd:\MLP{} performs poorly, \mcd:\CAT{} outperforms all others by a wide margin. This indicates that this time, \MLP{} and \CAT{} are complementary, as together they can obtain at least Top~ 2 performances in all cases besides the WineRed and WineWhite datasets.
\end{itemize}
\begin{table}[http]
\centering
\scriptsize
\begin{tabular}{|l||c|c|c|c|c|c|c|}
\hline
\vspace{-0.22cm}
 & & & & & & & \\
\multirow{2}{*}{Empirical \pNLL} & \mcd & \mcd& \multirow{2}{*}{\nnk} & \flex & \flex & \multirow{2}{*}{\kmn} & \multirow{2}{*}{\nf} \\ 
& \MLP & \CAT &  & \NeurN & \XGB &  &  \\
\hline
\hline
\vspace{-0.22cm}
 & & & & & & &  \\
BostonHousing & \underline{-0.64} & \textbf{-0.59} & -0.81 & -1.99 & -1.84 & -1.22 & -1.63 \\
Concrete & \textbf{-0.86} & \underline{-1.02} & -1.23 & -2.26 & -1.25 & -2.30 & -2.13 \\
NCYTaxiDropoff:lon. & \underline{-1.30} & \textbf{-1.28} & -1.68 & -2.05 & -2.17 & -2.51 & -2.85 \\
NCYTaxiDropoff:lat. & \textbf{-1.31} & \textbf{-1.31} & \underline{-1.44} & -1.67 & -6.18 & -2.45 & -2.96 \\
Power & \textbf{-0.06} & -0.36 & -0.73 & -1.09 & -0.75 & \underline{-0.35} & -0.39 \\
Protein & \textbf{-0.09} & \underline{-0.42} & -0.77 & -0.83 & -1.38 & -0.68 & -0.54 \\
WineRed & -0.89 & -0.89 & \textbf{3.486} & \underline{1.062} & 0.965 & -0.90 & -2.43 \\
WineWhite & -1.18 & -1.13 & \textbf{2.99} & -0.73 & \underline{-0.63} & -1.7 & -4.18 \\
Yacht & 0.14 & \textbf{0.822} & -0.46 & -1.23 & 0.144 & 0.025 & \underline{0.401} \\
teddy & \textbf{-0.47} & \underline{-0.51} & -0.83 & -0.83 & -1.34 & -0.76 & -0.94 \\
toy dataset 1 & -0.99 & \underline{-0.47} & -0.63 & -0.88 & -1.46 & \textbf{-0.35} & -0.71 \\
toy dataset 2 & -1.40 & \underline{-1.33} & \textbf{-1.31} & -1.54 & -1.43 & \underline{-1.33} & -1.39 \\
\hline
\end{tabular}
\caption{Evaluation of the negative log-likelihood (\pNLL) for $12$ datasets. \textbf{Best performance} is in bold print, \underline{second best performance} is underlined. Higher values are better. Methods included outperform all others besides \mcd on at least one dataset.} 
\label{table:real}
\end{table}

%%%%%%%%%%%%%%%%%%%%%%%%%%%%%%%%%%%%%%%%%%%%%%%%
\section{Ablation}\label{Sec:ablation}
%%%%%%%%%%%%%%%%%%%%%%%%%%%%%%%%%%%%%%%%%%%%%%%%
We now present an ablation study of the impact of the construction strategy used to build $\mDwz{\SampSizewz}$ and the chosen value for ratio $\ratio$. We also assess the impact of additional data on performances in Framework~\ref{DATA2} and ~\ref{DATA3}. Our experiments are done on the \dAsym density model, which corresponds to a case where \mcd{} is performing well but there is still room for improvement.\\

\begin{itemize}
    \item[$\bullet$] The main take away is that in Framework~\ref{DATA1}, \pUID-Construction~\ref{pro:ConsUID} is far better than \pIID-Construction~\ref{pro:ConsIID}.
    \item[$\bullet$] The appropriate ratio $\ratio$ for the \pUID-Construction~\ref{pro:ConsUID} should be around $0.05$, meaning $\SampSizewz = 20 \times \SampSizexy$.
    \item[$\bullet$] In Framework~\ref{DATA3}, as soon as two target values are associated to each observation, the \pUID-Construction~\ref{pro:ConsUIDMT} can massively improve the performances of \mcd, which are already very good when using \pUID-Construction~\ref{pro:ConsUID}.
\end{itemize}

% \noindent The main take away is that \pUID-Construction~\ref{pro:ConsUID} is far better than \pIID-Construction~\ref{pro:ConsIID}.  
%%%%%%%%%%%%%%%%%%%%%%%%%%%%%%%%%%%%%%%%%%%%%%%%
\paragraph{Construction strategy and ratio $\ratio$}
%%%%%%%%%%%%%%%%%%%%%%%%%%%%%%%%%%%%%%%%%%%%%%%%
\noindent\\
\noindent \ding{226} Table~\ref{table:ratio} depicts the impact of the construction strategy and ratio $\ratio$ on the performance in terms of empirical \pKL divergence in the classical Framework~\ref{DATA1} on the \dAsym{} density model. 
\begin{itemize}
\item  The appropriate ratio \ratio\, for the \pIID-Construction~\ref{pro:ConsIID} is $0.5$ which corresponds to a balanced distribution between the two classes.
    \item It seems clear that the \pUID-Construction~\ref{pro:ConsUID} produces much better results than the \pIID-Construction~\ref{pro:ConsIID}. For the latter, the performances are worse than those obtained with the concurrent method \nnk (see Table~\ref{table:density}: \pKL=0.245). On the other hand, the \pUID-Construction~\ref{pro:ConsUID} obtains Top~1 performances on our benchmark, by a wide margin.
    \item For the \pUID-Construction~\ref{pro:ConsUID}, it seems preferable to choose a ratio of $0.15$ or $0.05$, which corresponds respectively to a $6$ or $20$ times larger dataset. Indeed, in that case since $\SampSizewz = \frac{\SampSizexy}{\ratio}$, we have to make a trade-off between the size of the training dataset and the imbalance between the classes.
    \item Following these findings, in our benchmark, we choose to use the \pUID-Construction~\ref{pro:ConsUID} with a ratio of $0.05$.
\end{itemize}
%Il parait clair que la construction uid produit de bien meilleurs résultats que la construction iid. Pour la construction iid, les performances sont moins bonnes que celles obtenues avec la méthode concurrente \nnk ou avec le discriminateur Catboost. Le ratio approprié dans ce cas-ci est de 0.5 qui correspond à une répartition équilibrée entre les deux classes. La construction uid en revanche permet d'obtenir des résultats exceptionnels. Dans ce cas-là, pour choisir le ratio approprié il y a un arbitrage a faire entre la taille du jeu de donnée et le déséquilibre entre classes. On voit qu'on a intérêt à choisir un ratio de 0.15 ou 0.05 ce qui correspond à des jeux de données 6 et 20 fois supérieurs respectivements. Dans notre benchmark, on a choisit d'utiliser la Construction uid avec un ratio de 0.05.
\begin{table}[http]
\centering
\scriptsize
\begin{tabular}{|l||c|c||c|}
\hline
\vspace{-0.25cm}
 & & & \\
Construction & Ratio $\ratio$ & $\SampSizewz$ & $\pKL$ \\
\hline
\hline
\vspace{-0.25cm}
 & & & \\
\multirow{4}{*}{$\pIID$-Construction~\ref{const:DwzNiid}} & 0.05 & 50 & 0.3284 \\
 & 0.15 & 50 & 0.2865 \\
 & 0.5 & 50 & \textbf{0.2771} \\
 & 0.85 & 50 & 0.4211 \\

\hline
\vspace{-0.25cm}
 & & & \\
 \multirow{5}{*}{$\pUID$-Construction~\ref{const:DwzNid}} & 0.01 & 10000 & 0.0563 \\
 & 0.015 & 6666 & \textbf{0.0546} \\
 & 0.05 & 2000 & 0.0550 \\
 & 0.15 & 666 & 0.0551 \\
 & 0.5 & 200 & 0.0996 \\
\hline
\end{tabular}
\caption{Evaluation of the \pKL divergence of the \mcd:\MLP method on the \dAsym density model in Framework~\ref{DATA1} with various values for ratio $\ratio$, with $\pIID$ and $\pUID$ constructions. \textbf{Best} performance for each construction is in bold print.}
\label{table:ratio}
\end{table}

%There is not one strategy which outperforms all others in all cases. It is clear however, that larger ratios (and consequently smaller training set sizes) can lead to numerical instability. Note that when using the $\pIID$ construction~\ref{pro:ConsIID}, the dataset size is divided by two, while for the construction~\ref{pro:ConsUID}, we have $\SampSizewz= \frac{\SampSizexy}{\ratio}$, meaning smaller ratios correspond to larger training sets. For the \dJump, \dLinG and \dStud density models, the construction strategy does not have a significant impact on the performance. Meanwhile, in the other cases the results vary significantly. Interestingly, the $\pIID$ property is not always desirable, as it leads to converge issues in $3$ cases, and construction~\ref{pro:ConsIID} under-performs by a large margin when compared with the construction~\ref{pro:ConsUID} with the same $\ratio=0.5$ in two other cases. The $\pIID$ construction~\ref{pro:ConsIID} did outperform all others on \dSkew, with a \pKL of $0.545$ but this is still far worst then \mdn, the best performing method, which obtains a $\pKL = 0.014$. The two reasonable choices in terms of ratio values seem to be $\ratio=0.5$ and $\ratio=0.05$. In the other experiments, we always used the construction~\ref{pro:ConsUID} with $\ratio=0.05$.
%%%%%%%%%%%%%%%%%%%%%%%%%%%%%%%%%%%%%%%%%%%%%%%%
\paragraph{Additional marginal data}
%%%%%%%%%%%%%%%%%%%%%%%%%%%%%%%%%%%%%%%%%%%%%%%%
\noindent\\
\noindent \ding{226} Table~\ref{table:marginal} depicts the impact on $\pKL$ performance when using Constructions~\ref{const:DwzNiidAD} and ~\ref{const:DwzNidAD} (corresponding to the \pIID{} and \pUID{} case respectively) in Framework~\ref{DATA2} where additional marginal data is available. We compare the respective benefit of adding only marginal observations ($\SampSizex > 0,\, \SampSizey = 0$), only marginal target values ($\SampSizex = 0,\, \SampSizey > 0$), and both marginal observation and marginal target values ($\SampSizex > 0, \SampSizey > 0$).\\

\begin{itemize}
    \item In the \pIID\,case, using \pIID-Construction~\ref{const:DwzNiidAD} instead of \pIID-Construction~\ref{const:DwzNiid} produces substantial improvements in terms of performances,  but not enough to outperform the \pUID-Construction~\ref{const:DwzNid}. When using the \pUID-Construction~\ref{const:DwzNidAD} instead of the \pUID-Construction~\ref{const:DwzNid}, the performance gain is smaller, but bear in mind that performances are already very satisfying at that point.
    \item In the \pIID\,case, having access to $\SampSizey = \SampSizexy$ marginal target values also allows the marginal estimator to be trained on a sample size of $\SampSizey + \SampSizexy = 2 \SampSizexy$, which may explain why the performance gain is larger with marginal target values $\mDy$ than with marginal observations $\mDx$ when using the \pIID-Constructions~\ref{const:DwzNiidAD}.
    \item Meanwhile, in the \pUID case, the size of the training dataset $\mDwz{\SampSizewz}$ when $\max(\SampSizex,\SampSizey)> 0$ is at most $(\SampSizexy+ \SampSizex)(\SampSizexy+\SampSizey)> \SampSizexy^2 $, which allows to build an even larger data set ($\SampSizewz > \SampSizexy^2$). However, in that case, the choice of $\SampSizewz$ is constrained by the choice of ratio $\ratio$: $\SampSizewz = \frac{\SampSizexy}{\ratio}$. Here the appropriate ratio is $\ratio = 0.05$ and $\SampSizexy = 100$, meaning $\SampSizewz = 2000 \leq \SampSizexy^2$. As such, using Construction~\ref{pro:ConsUIDAM} to increase $\SampSizewz$ beyond $\SampSizexy^2$ is not useful in that setting.
    \item The Construction~\ref{pro:ConsUIDAM} does, however, increase the amount of information present in the dataset $\mDwz{\SampSizewz}$. Here, the performance gain is higher in the presence of marginal observations $\mDx$ than marginal target values $\mDy$. This may be partly due to the fact that in the \dAsym model, $\vX \in \eR^{10}$ and $\vY \in \eR$, meaning the observations contain more information than the target values.
\end{itemize}

\begin{table}[http]
\centering
\footnotesize
\begin{tabular}{|l|c|c|c||c|c|c||c|}
\hline
\multicolumn{4}{|c||}{Setting}&\multicolumn{3}{c||}{Construction} & Results\\
\hline
\vspace{-0.22cm}
 & & & & & & & \\
Available datasets &  $\SampSizexy$ & $\SampSizex$ & $\SampSizey$ & Strategy & Construction &  $\SampSizewz$ & \pKL \\
\hline
\hline
\vspace{-0.22cm}
 & & & & & & & \\
\mDxy & 100 & 0 & 0 & & Construction~\ref{const:DwzNiid}  & 50  &  0.3456 \\
\mDxy, \mDx  & 100 & 100 & 0  & \pIID & Construction~\ref{const:DwzNiidAD}  & 100 &  \textbf{0.1204} \\
\mDxy, \mDy &   100 & 0 & 100 & $\ratio = 0.5$ &  Construction~\ref{const:DwzNiidAD}  & 100 &  \textbf{0.1203} \\
\mDxy, \mDx , \mDy &  100 & 25 & 25 &  & Construction~\ref{const:DwzNiidAD}  & 75 &  0.2058 \\
\hline
\vspace{-0.22cm}
 & & & & & & & \\
\mDxy & 100 & 0 & 0 &  & Construction~\ref{const:DwzNid}  & 2000 & 0.0551 \\
\mDxy, \mDx & 100 & 500 & 0 & \pUID & Construction~\ref{const:DwzNidAD}  & 2000 & \textbf{0.0541} \\
\mDxy, \mDy & 100 & 0 & 500 & $\ratio = 0.05$ & Construction~\ref{const:DwzNidAD}  & 2000 & 0.0546 \\
\mDxy, \mDx , \mDy & 100 & 150 & 150 &  & Construction~\ref{const:DwzNidAD}  & 2000 & 0.0639 \\
\hline
\end{tabular}
\caption{Evaluation of the \pKL divergence values of the \mcd:\MLP method on the \dAsym density model in Framework~\ref{DATA1} and ~\ref{DATA2} with various values for $\SampSizexy$, $\SampSizex$ and $\SampSizey$, with $\pIID$ and $\pUID$ constructions. \textbf{Best} performance for $\pIID$ and $\pUID$ are in bold print.}
\label{table:marginal}
\end{table}
\paragraph{Multiple target values per observations}
%%%%%%%%%%%%%%%%%%%%%%%%%%%%%%%%%%%%%%%%%%%%%%%%
\noindent\\
\noindent \ding{226} Table~\ref{table:multisample} compares the $\pKL$ performances of \pUID-Construction~\ref{const:DwzNidFram3} in Framework~\ref{DATA3}, when more than one target is associated to each observation in the dataset $\mDxyM$. Here we denote $\SampSizeM$ the number of observations associated to each target. Remark that when $\SampSizeM = 1$, \pUID-Construction~\ref{const:DwzNidFram3} is strictly equivalent to \pUID-Construction~\ref{const:DwzNid}.\\

\begin{itemize}
    \item In the presence of multiple target values per observation, \pUID-Construction~\ref{const:DwzNidFram3} allows for unparalleled performances. This can probably be explained by the fact that the goal of the \cde task is to determine the relationship between $\vX$ and $\vY$ beyond the conditional expectation, and thus multiple realizations for a single observation better quantify the variance.  
    \item Besides, it seems that when $\SampSizeM > 1$, the appropriate ratio is higher, since the performances for $\ratio = 0.15$ are better than with $\ratio=0.05$, which is not the case when $\SampSizeM = 1$.
\end{itemize}

\begin{table}[http]
\centering
\footnotesize
\begin{tabular}{|l||c|c|c||c|}
\hline
Construction & $\SampSizeM$ & Ratio $\ratio$ & $\SampSizewz$ & $\pKL$ \\
\hline
\hline
\vspace{-0.26cm}
 & & & & \\
 &  & 0.5 & 200 & 0.0620 \\
\pUID-Construction~\ref{const:DwzNid} & 1 & \underline{0.15} & 666 & 0.0502 \\
 &   & \underline{0.05} & 2000 & 0.0501 \\
\hline
\vspace{-0.26cm}
 & & & & \\
 &   & 0.5 & 400 & 0.0520 \\
\pUID-Construction~\ref{const:DwzNidFram3} & 2 & \underline{0.15} & 1333 & 0.0438 \\
 &   & 0.05 & 4000 & 0.0461 \\
\hline
\vspace{-0.26cm}
 & & & & \\
 &   & 0.5 & 2000 & 0.0179 \\
\pUID-Construction~\ref{const:DwzNidFram3} & 10 &  \underline{0.15} & 6666 & \textbf{0.0167} \\
 &   & 0.05 & 20000 & 0.0191 \\
\hline
\end{tabular}
\caption{Evaluation of the \pKL divergence values of the \mcd:\MLP method on the \dAsym density model in Framework~\ref{DATA1} and ~\ref{DATA3} with various values for ratio $\ratio$ and $\SampSizeM$. \textbf{Best} performance is in bold print. \underline{Best} performance ratio for each value of $\SampSizeM$ is underlined. Column 2 corresponds to the number of target values associated to each observation.}
\label{table:multisample}
\end{table}

%% file: Proofs.tex
\section{Proofs and dataset constructions}\label{Sec:proofs}
%%%%%%%%%%%%%%%%%%%%%%%%%%%%%%%%%%%%%%%%%%%

%%%%%%%%%%%%%%%%%%%%%%%%%%%%%%%%%%%%%%%%%%%
%\subsection{Theoretical properties}
%%%%%%%%%%%%%%%%%%%%%%%%%%%%%%%%%%%%%%%%%%%

%%%%%%%%%%%%%%%%%%%%%%%%%%%%%%%%%%%%%%%%%%%
\subsection{Proof Fact~\ref{fac:MCF}}
%%%%%%%%%%%%%%%%%%%%%%%%%%%%%%%%%%%%%%%%%%%

Let $(x,y)\in \mathcal{X}\times \mathcal{Y}$ be any couple of values for which we need to prove Fact~\ref{fac:MCF}. %By abuse of notation we denote $\pf{\vX}(x) = \pf{\vX}$ and $\pf{\vY}(y) = \pf{\vY}$. 
We have by the $Bayes$'s formula:
\begin{equation}
    \frac{\pf{\vY\pCond \vX=x}(y)}{\pf{\vY}(y)} = \frac{\pf{\vX,\vY}(x,y)}{\pf{\vX}(x)\pf{\vY}(y)},\,  \pf{\vX}(x)> 0, \, \pf{\vY}(y)>0 \label{eq:Bayes}
\end{equation}

Then, for any $\ratio \in (0,1)$:
$$
\frac{\pf{\vX,\vY}(x,y)}{\pf{\vX}(x)\pf{\vY}(y)} = \frac{\ratio\pf{\vX,\vY}(x,y)}{\ratio\pf{\vX,\vY}(x,y) + (1 - \ratio)\pf{\vX}(x)\pf{\vY}(y)}\times  \frac{\ratio\pf{\vX,\vY}(x,y) + (1 - \ratio)\pf{\vX}(x)\pf{\vY}(y)}{\ratio \pf{\vX}(x)\pf{\vY}(y)}
$$
Let $\contrast(x,y) := \frac{\ratio\pf{\vX,\vY}(x,y)}{\ratio\pf{\vX, \vY}(x,y) + (1 - \ratio)\pf{\vX}(x)\pf{\vY}(y)}$ the marginal constrast function with ratio $\ratio$ defined on Definition~\ref{def:MCF}
\begin{align*}
\frac{\pf{\vX,\vY}(x,y)}{\pf{\vX}(x)\pf{\vY}(y)}&= \contrast(x,y)\times  \left(\frac{\pf{\vX,\vY}(x,y)}{\pf{\vX}(x)\pf{\vY}(y)} +\frac{1 - \ratio}{\ratio}\right)\\
\Leftrightarrow
(1-\contrast(x,y))\frac{\pf{\vX,\vY}(x,y)}{\pf{\vX}(x)\pf{\vY}(y)}&= \contrast(x,y)\frac{1 - \ratio}{\ratio} \\
\Leftrightarrow\frac{\pf{\vY\pCond \vX=x}(y)}{\pf{\vY}(y)} &= \frac{1-\ratio}{\ratio}\frac{\contrast(x,y)}{1-\contrast(x,y)}
\end{align*}
The last equation is deduced using equation~\eqref{eq:Bayes}. 

%%%%%%%%%%%%%%%%%%%%%%%%%%%%%%%%%%%%%%%%%%%%%
\subsection{Proof Proposition~\ref{pro:equiv}}\label{sec:Proofequiv}
%%%%%%%%%%%%%%%%%%%%%%%%%%%%%%%%%%%%%%%%%%%%%

By condition~\ref{MDC1} of Definition~\ref{def:MDC} we have $\vZ \sim \pBB(r)$, then 
$$
\pP[\vZ = 1] = r\ \text{ and}  \ \pf{\vZ }(z)=p^{z}(1-p)^{1-z}.
$$
Moreover, by condition~\ref{MDC3} of Definition~\ref{def:MDC}, we have $\forall (x,y) \in \pOmegX \times \pOmegY$, $\pf{\vW }(x,y) = r\pf{\vX,\vY}(x,y)+ (1-r)\pf{\vX}(x)\pf{\vY}(y)$. %By abuse of notation we denote $\pf{\vX}(x) = \pf{\vX}$, $\pf{\vY}(y) = \pf{\vY}$, $\pf{\vW }(x,y) = \pf{\vW }$ and $\pf{\vZ }(z) = \pf{\vZ }$.
By condition~\ref{MDC4} of Definition~\ref{def:MDC} we have 
\begin{align*}
    \pf{\vW \pCond \vZ = 1}(x,y) &= \pf{\vX,\vY}(x,y)\\
    \pf{\vW \pCond \vZ = 0}(x,y) &= \pf{\vX}(x)\pf{\vY}(y)
\end{align*}
By Definition~\ref{def:MCF} we have 
\begin{align*}
    \contrast(x,y) &= \frac{\ratio \times \pf{\vX,\vY}(x,y)}{\ratio \pf{\vX,\vY}(x,y) + (1-\ratio)\pf{\vX}(x)\pf{\vY}(y)} 
    = \frac{\pP[\vZ = 1] \times \pf{\vW \pCond \vZ = 1}(x,y)}{\ratio \pf{\vW \pCond \vZ = 1}(x,y) + (1-\ratio)\pf{\vW \pCond \vZ = 0}(x,y)}\\
     &= \frac{\pP[\vZ = 1] \times \pf{\vW \pCond \vZ = 1}(x,y)}{\pf{\vW }(x,y)}
    = \frac{\pEz[\Indi_{\vZ = 1}] \times \left[1\times \pf{\vW \pCond \vZ = 1}(x,y)+0\times \pf{\vW \pCond \vZ = 0}(x,y)\right]}{\pf{\vW}(x,y)}\\
    &= \frac{\pEz[\Indi_{\vZ = 1}] \times \pEz[\pf{\vW \pCond \vZ=z}]}{\pf{\vW}(x,y)}
    = \frac{\pEz[\Indi_{\vZ = 1}\pf{\vW \pCond \vZ = z}(x,y)] }{\pf{\vW}(x,y)}
    = \frac{\pEz[\Indi_{\vZ = 1}\pf{\vW ,\vZ}(x,y,z)] }{\pf{\vW}(x,y)\pf{\vZ }(z)}\\
    &= \pEz\left[\Indi_{\vZ = 1}\times\frac{\pf{\vZ \pCond \vW = (x,y)}(z)}{\pf{\vZ }(z)}\right]
    = \pEz\left[\Indi_{\vZ = 1}\times\frac{\pP\left[\vZ = z\pCond \vW = (x,y)\right]}{\pP[\vZ = z]}\right]\\ 
    &= \pEz\left[\Indi_{\vZ = 1}\times\frac{\pP\left[\vZ = 1\pCond \vW = (x,y)\right]}{\pP[\vZ = 1]}\right]
    = \pEz[\Indi_{\vZ = 1}]\frac{\pP[\vZ = 1\pCond \vW = (x,y)]}{\pP[\vZ = 1]}\\ 
    &= \pP[\vZ = 1 \pCond \vW = (x,y)] = \pE[\vZ \pCond \vW = (x,y)]
\end{align*}

%%%%%%%%%%%%%%%%%%%%%%%%%%%%%%%%%%%%%%%%%%%%%
%\subsection{Proofs and Dataset Constructions}
%%%%%%%%%%%%%%%%%%%%%%%%%%%%%%%%%%%%%%%%%%%%%

%%%%%%%%%%%%%%%%%%%%%%%%%%%%%%%%%%%%%%%%%%%%%
\subsection{Proof Theorem~\ref{pro:ConsIID} and Construction in the i.i.d. case}
\label{Sec:ProofConsIID}
%%%%%%%%%%%%%%%%%%%%%%%%%%%%%%%%%%%%%%%%%%%%%
First construct a random vector $(\vW,\vZ)$ satisfying the $\MDC(\ratio)$ with $\ratio \in \pOmegr$.
\begin{itemize}
    \item Consider the random vector $(\vX,\vY)$ admitting $\pf{\vX,\vY}$ as density of probability.
    \item Let $\vYtild$ be a random variable independent of $\vX$ and $\vY$ and of same law $\pf{\vY}$ of $\vY$ ($\vYtild \pequal \vY$).
    \item Let $\ratio$ be a real number in $\pOmegr$ and $\vZ \sim \pBB(\ratio)$.
    \item Set $\vW = (\vX,\vY \Indi_{\vZ = 1} + \vYtild \Indi_{\vZ = 0})$.
\end{itemize}
Then, for all $(x,y,\ytild) \in \pOmegX \times \pOmegY \times \pOmegY$ we have $$\forall z \in \pOmegZ \, \pf{\vW \pCond \vZ = z}(x,y,\ytild) = \left\{
\begin{array}{ll}
 \pf{\vX ,\vY}(x,y) & \mbox{if } z = 1 \\
 \pf{\vX}(x)\pf{\vY}(\ytild) & \mbox{if } z = 0
\end{array}
\right.$$ 

\noindent Therefore, $\forall (x,y,z) \in \pOmegX \times \pOmegY \times \pOmegZ$, $\pf{\vW \pCond \vZ = z}(x,y) = \pf{\vX,\vY}(x,y) \Indi_{z = 1} + \pf{\vX}(x)\pf{\vY}(y) \Indi_{z = 0})$.

Moreover, $\forall (x,y) \in \pOmegX \times \pOmegY$
\begin{align*}\label{eq:4imp3lol3}
\pf{\vW}(x,y) &= \Integ{}{}\pf{\vW \pCond \vZ = z}(x,y) \pf{\vZ}(z) \, dz\\
    &=  \pf{\vW \pCond \vZ = 1}(x,y)\pP[\vZ = 1] + \pf{\vW \pCond \vZ = 0}(x,y)\pP[\vZ = 0]\\
    &= \pf{\vW \pCond \vZ = 1}(x,y)\ratio + \pf{\vW \pCond \vZ = 0}(x,y)(1 - \ratio) \\
    &=  \ratio\pf{\vX,\vY}(x,y)+ (1-\ratio)\pf{\vX}(x)\pf{\vY}(y).
\end{align*}
Therefore $\vZ,\vW$ satisfies Definition~\ref{def:MDC}. Now, consider the original $\SampSizexy$-sample $\mDxy$ and set $\SampSizewz = \ArrB{\SampSizexy/2}$. We can now construct the $\mDwz{\SampSizewz}$ sample.

\vspace{0.5cm}

%%%%%%%%%%%%%%%%%%%Construction 1%%%%%%%%%%%%%%%%%%%%%%%%%%%%%%%%%%%%%%%
\noindent
\begin{minipage}{5.5cm}
\begin{mdframed}
\vspace{-0.1cm}
\begin{const} 
\label{const:DwzNiid}
\big[$\pIID$ $\mDwz{\SampSizewz}$\big]
\end{const}
\vspace{-0.1cm}
\end{mdframed}
\end{minipage}
\vspace{0.5cm}\\
Consider the original $\SampSizexy$-sample $\mDxy$ and set $\SampSizewz = \ArrB{\SampSizexy/2}$. We can now construct the $\mDwz{\SampSizewz}$ sample:\\

\noindent
\STEPUIDOne\label{stepone}: \quad \begin{tabular}[t]{|l}
\begin{minipage}[t]{10cm}
First sample $\SampSizewz$ independent observations $\vZ_1, \cdots , \vZ_{\SampSizewz}$ with respect to $\pBB(\ratio)$.
\end{minipage}
\end{tabular}
\vspace{0.5cm}

\noindent\STEPUIDTwo\label{steptwo}: \quad \begin{tabular}[t]{|l}
\begin{minipage}[t]{10cm}
Next, $\forall i = 1, \cdots , \SampSizewz$, set $ \vW_i = (\vX_i, \vY_i \Indi_{\vZ_i = 1} + \vY_{i + \SampSizewz} \Indi_{\vZ_i = 0})$.
\end{minipage}
\end{tabular}\\

Since $\SampSizewz>0$ and the $\SampSizewz$ couples $(\vW_i, \vZ_i)_i$ are constructed from the $\SampSizewz$-$\pIID$ quadruplets $(\vX_i, \vY_i, \vY_{i + \SampSizewz}, \vZ_i)_i$, they are $\pIID$.

%%%%%%%%%%%%%%%%%%%%%%%%%%%%%%%%%%%%%%%%%%%%%
\subsection{Proof Theorem~\ref{pro:ConsUID} and Construction in the i.d. case}
\label{Sec:ProofConsUID}
%%%%%%%%%%%%%%%%%%%%%%%%%%%%%%%%%%%%%%%%%%%%%

Let $n_J$ and $n_M$ two integers such that $1 \leq n_J \leq \SampSizexy$ and $1 \leq n_M \leq \SampSizexy(\SampSizexy-1)$.

\vspace{0.5cm}

%%%%%%%%%%%%%%%%%%%Construction 2%%%%%%%%%%%%%%%%%%%%%%%%%%%%%%%%%%%%%%%
\noindent
\begin{minipage}{5.5cm}
\begin{mdframed}
\vspace{-0.1cm}
\begin{const}
\label{const:DwzNid}
\big[$\pUID$ $\mDwz{\SampSizewz}$\big]
\end{const}
\vspace{-0.1cm}
\end{mdframed}
\end{minipage}
\vspace{0.5cm}

\noindent
\STEPUIDOne\label{stepone2}: \quad \begin{tabular}[t]{|l}
\begin{minipage}[t]{10cm}
Construct $\SampSizexy^2$ observations $\{(\vWtild_i,\vZtild_i)\}_{i = 1, \cdots, \SampSizexy^2}$ such that:
$\forall j = 0, \cdots, \SampSizexy - 1, \, \forall k = 1, \cdots, \SampSizexy \, : \,$
\begin{equation}\label{eq:steponelol}
\left\{
\begin{array}{ll}
\vWtild_{j \SampSizexy + k} = (\vX_{j+1}, \vY_k) \\
\vZtild_{j \SampSizexy + k} = \Indi_{j+1 = k}
\end{array}
\right.
\end{equation}
\end{minipage}
\end{tabular}

\noindent Note that the $\SampSizexy^2$ observations are neither independent nor identically distributed and do not satisfy yet the $\MDC$.\\

\noindent
\STEPUIDTwo\label{steptwo2}: \quad \begin{tabular}[t]{|l}
\begin{minipage}[t]{10cm}
Split the $\SampSizexy^2$ couple of observations $\{(\vWtild_i,\vZtild_i)\}_{i = 1, \cdots, \SampSizexy^2}$ into two:
$$
\left\{
\begin{array}{ll}
S_J \,= \{(\vWtild_i,\vZtild_i)\} : \, \vZtild_i = 1 \, \forall i  \\
S_M = \{(\vWtild_i,\vZtild_i)\} : \, \vZtild_i = 0 \, \forall i .
\end{array}
\right.
$$
\end{minipage}
\end{tabular}

\vspace{0.5cm}
\noindent
\STEPUIDThree\label{stepthree2}: \quad \begin{tabular}[t]{|l}
\begin{minipage}[t]{10cm}
Sample at random uniformly without replacement $n_J$ observations $(\vWtild_i,\vZtild_i)$ from $S_J$ and denote $\widetilde{S}_J$ this $n_J$ sample.
\end{minipage}
\end{tabular}\\

\noindent Note that since $\vZtild_i = 1$ if and only if $j+1 = k$ in equation\eqref{eq:steponelol}, we have
\begin{equation}\label{eq:SJlol1}
    \forall (\vWtild_i,\vZtild_i) \in \widetilde{S}_J, \vZtild_i = 1 \text{ and }\pf{\vWtild_i \pCond \vZtild_i = 1} \equiv \pf{\vX,\vY} 
\end{equation}
This means that $\vWtild_{j(\SampSizexy + 1) + 1} = (\vX_{j+1},\vY_{j+1})$.

\vspace{0.5cm}
\noindent
\STEPUIDFour\label{stepfour2}: \quad \begin{tabular}[t]{|l}
\begin{minipage}[t]{10cm}
Sample at random uniformly without replacement $n_M$ observations $(\vWtild_i,\vZtild_i)$ from $S_M$ and denote $\widetilde{S}_M$ this $n_M$ sample.
\end{minipage}
\end{tabular}\\

\noindent Note that since $\vZtild_i = 0$ if and only if $j+1 \neq k$ in equation\eqref{eq:steponelol}, we have
\begin{equation}\label{eq:SJlol2}
 \forall (\vWtild_i,\vZtild_i) \in \widetilde{S}_M, \vZtild_i = 0 \text{ and }\pf{\vWtild_i \pCond \vZtild_i = 0} \equiv \pf{\vX}\pf{\vY} 
\end{equation}
This means that $\vWtild_{j\SampSizexy + k} = (\vX_{j+1},\vY_{k})$ (recall $\vX_l \pIndep \vX_k \, \forall l \neq k$).

\vspace{0.5cm}
\noindent
\STEPUIDFive\label{stepfive2}: \quad \begin{tabular}[t]{|l}
\begin{minipage}[t]{10cm}
Concatenate the samples $\widetilde{S}_J$ and $\widetilde{S}_M$ and shuffle uniformly to obtain a $\SampSizewz$ sample $\mDwz{\SampSizewz}= \mathcal{U}\texttt{nif.}\mathcal{S}\texttt{huffle} \left(\{ \widetilde{S}_J; \widetilde{S}_M\}\right)$.
\end{minipage}
\end{tabular}

\vspace{0.5cm}
\noindent Prove now that $\forall i = 1, \cdots , \SampSizewz,$ the couple $(\vW_i, \vZ_i) \in \mDwz{\SampSizewz}$ satisfies $\MDC(\frac{n_J}{\SampSizewz})$.
\begin{itemize}
\item[\eqref{MDC1}] As we shuffle uniformly the indices, we have $\forall i = 1, \cdots, \SampSizewz, \, \vZ_i \in \mDwz{\SampSizewz}$, $\vZ_i \sim \pBB(\frac{n_J}{\SampSizewz})$.
\item[\eqref{MDC2}]  By \STEPUIDOne, it is obvious that all the $\vW_i$ admit $\pOmegX \times \pOmegY$ as support.
\item[\eqref{MDC4}]  Let $(\vW_i, \vZ_i)$ be any element of $\mDwz{\SampSizewz}$, then by  equation~\eqref{eq:SJlol1}
$$\left\{
\begin{array}{ll}
 \pf{\vW_i \pCond \vZ_i = 1} \equiv \pf{\vX,\vY} &\mbox{if } \vZ_i = 1 \\
\pf{\vW_i \pCond \vZ_i = 0} \equiv \pf{\vX}\pf{\vY} &\mbox{if } \vZ_i = 0 
\end{array}
\right.$$
\item[\eqref{MDC3}]  Moreover, it  comes $\pf{\vW_i} (x,y) = \ratio \pf{\vX, \vY}(x,y) + (1 - \ratio) \pf{\vX}\pf{\vY}$ with $\ratio = \frac{n_J}{\SampSizewz}$.
\end{itemize}

%%%%%%%%%%%%%%%%%%%%%%%%%%%%%%%%%%%%%%%%%%
\subsection{Proof Theorem \ref{pro:ConsIIDAM} and i.i.d. Construction in the Additional data setting}
\label{Sec:ProofConsIIDAM}
%%%%%%%%%%%%%%%%%%%%%%%%%%%%%%%%%%%%%%%%%%%%%

To construct $\mDwz{\SampSizewz}= \{(\vW_i,\vZ_i)\}_{i = 1, \cdots, \SampSizewz}$ a training set of $\SampSizewz$ $\pIID$ observations, we concatenate $3$ datasets denoted $\mDwzRes{\SampSizewzx}$, $\mDwzRes{\SampSizewzy}$ and $\mDwzRes{\SampSizewzxy}$ of respective size $\SampSizewzx$, $\SampSizewzy$ and $\SampSizewzxy$ such that
$$\mDwz{\SampSizewz} = \mDwzRes{\SampSizewzx} \cup \mDwzRes{\SampSizewzy} \cup \mDwzRes{\SampSizewzxy}
$$
$$\text{and} \ 
\left\{\begin{array}{ll}
 \SampSizewzx = \min(\SampSizexy, \SampSizex),   &\SampSizewzy = \min(\SampSizey, \SampSizexy - \SampSizewzx) \\
  \SampSizewzxy = \ArrB{\frac{\SampSizexy - \SampSizewzx - \SampSizewzy}{2}},  & \SampSizewz = \SampSizewzx + \SampSizewzy + \SampSizewzxy
\end{array}
\right.
$$
To construct these $3$ preliminary datasets, we first split $$\mDxy= \underbrace{\mDxyRes{2\SampSizewzxy}}_{\text{the first $ 2\SampSizewzxy$ obs.}} \cup \underbrace{\mDxyRes{\SampSizewzy}}_{\text{the next $\SampSizewzy$ obs.}} \cup \underbrace{\mDxyRes{\SampSizewzx}}_{\text{the next $\SampSizewzx$ obs.}} \cup \underbrace{\mathcal{D}}_{\text{the rest.}}
$$

\noindent We consider $\mDxRes{\SampSizewzx}$ and $\mDyRes{\SampSizewzy}$ the observations in datasets $\mDx$ and $\mDy$ restricted to the $\SampSizewzx$ and $\SampSizewzy$ first observations respectively.

\vspace{0.5cm}

%%%%%%%%%%%%%%%%%%%Construction 3%%%%%%%%%%%%%%%%%%%%%%%%%%%%%%%%%%%%%%%
\noindent
\begin{minipage}{8.5cm}
\begin{mdframed}
\vspace{-0.1cm}
\begin{const}
\label{const:DwzNiidAD}
\big[$\pIID$ $\mDwz{\SampSizewz}$ Additional data\big]
\end{const}
\vspace{-0.1cm}
\end{mdframed}
\end{minipage}

\vspace{0.5cm}
\noindent\STEPUIDOne\label{stepone3}: \quad \begin{tabular}[t]{|l}
\begin{minipage}[t]{10cm}
\textbf{Construction of } $\mDwzRes{\SampSizewzxy}$. If $\SampSizewzxy = 0$, then $\mDwzRes{\SampSizewzxy} = \emptyset$, otherwise we construct $\mDwzRes{\SampSizewzxy}$ from the initial dataset $\mDxRes{\SampSizewzx}$ as described in Construction~\ref{const:DwzNiid}.

\end{minipage}
\end{tabular}

\vspace{0.5cm}

\noindent\STEPUIDTwo\label{steptwo3}: \quad \begin{tabular}[t]{|l}
\begin{minipage}[t]{10cm}
\textbf{Construction of } $\mDwzRes{\SampSizewzy}$.First note that if $\SampSizewzy = 0$, then $\mDwzRes{\SampSizewzy}= \emptyset$. If $\SampSizewzy \neq 0$, we proceed as follows:\\

\begin{tabular}[t]{|c}\hspace{-0.5cm}
\begin{minipage}{12cm}
\begin{enumerate}[label=(\roman*)]
    \item Sample $\SampSizewzy$ independent observations $(\vZ_i)_{i = 1, \cdots, \SampSizewzy}$ according to a Bernoulli law of parameter $\ratio \in (0,1)$.
    \item For all $i = 1, \cdots, \SampSizewzy$; $\forall (\vX_i, \vY_i) \in \mDxyRes{\SampSizewzy}$ and $\forall \vYtild_i \in \mDyRes{\SampSizewzy}$, set $$\vW_i = (\vX_i, \vY_{i} \Indi_{\vZ_{i} = 1} + \vYtild_{i} \Indi_{\vZ_{ i} = 0}).$$
\end{enumerate}
\end{minipage}
\end{tabular}\\
\end{minipage}
\end{tabular}\\

\noindent Note that by construction, $(\vW_i, \vZ_{i})_{i=1}^{\SampSizewzy}$ are $\pIID$ and follow $\pf{\vW,\vZ}$. Moreover, following the same arguments as in proof of Theorem~\ref{pro:ConsIID}, the $\{ (\vW_i, \vZ_{i})\}_{i=1}^{\SampSizewzy}$ satisfy the $\MDC(\ratio)$.

\vspace{0.5cm}
\noindent\STEPUIDThree\label{stepthree3}: \quad \begin{tabular}[t]{|l}
\begin{minipage}[t]{10cm}
\textbf{Construction of } $\mDwzRes{\SampSizewzx}$.\\
\begin{tabular}[t]{|c}\hspace{-0.5cm}
\begin{minipage}{12cm}
\begin{enumerate}[label=(\roman*)]
\item Sample $\SampSizewzx$ independent observations $(\vZ_i)_{i = 1, \cdots, \SampSizewzx}$ according to a Bernoulli law of parameter $\ratio \in (0,1)$.
\item For all $i = 1, \cdots, \SampSizewzx$; $\forall (\vX_i, \vY_i) \in \mDxyRes{\SampSizewzx}$ and $\forall \vXtild_i \in \mDxRes{\SampSizewzx}$, set $$\vW_i= (\vX_i \Indi_{z = 1}+ \vXtild_i \Indi_{z = 0}, \vY_i)$$.
\end{enumerate}
\end{minipage}
\end{tabular}

\end{minipage}
\end{tabular}\\

\noindent Note that, by construction $\mDwzRes{\SampSizewzx} = (\vW_i, \vZ_{i})_{i=1}^{\SampSizewzx}$ are $\pIID$ and follow $\pf{\vW,\vZ}$. Indeed, the reasoning is similar as  in proof of Theorem~\ref{pro:ConsIID}. First construct a random vector $(\vW,\vZ)$ satisfying the $\MDC(\ratio)$ with $\ratio \in \pOmegr$: 
\begin{itemize}
    \item Let $(\vX,\vY)$ be a random variable admitting $\pf{\vX,\vY}$ as probability density function.
    \item Let $\vZ$ be a random variable following a Bernoulli of parameter $\ratio \in (0,1)$.
    \item Let $\vXtild$ be a random variable independent of $(\vX,\vY)$ following the law $\pf{\vX}$.
    \item Set $\vW = (\vX_{i} \Indi_{\vZ_{i} = 1} + \vXtild_{i} \Indi_{\vZ_{i} = 0}, \vY_i)$.
\end{itemize}
Then, $\forall (x, \widetilde{x},y) \in \pOmegX \times \pOmegY \times \pOmegY$ we have $\forall z \in \{0,1\}$,
$$
\pg{\vW \pCond \vZ = z}(x, \widetilde{x},y) = \left\{
\begin{array}{ll}
 \pf{\vX , \vY}(x,y) \text{ if }z = 1 \\
 \pf{\vX}(\widetilde{x})\pf{\vY}(y) \text{ if }z = 0
\end{array}
\right.$$\\
Then $\forall (x, y, z) \in \pOmegX \times \pOmegY \times \{0,1\}$, it comes $$\pf{\vW \pCond \vZ = z}(x, y) = \pg{\vW \pCond \vZ = z}(x, x,y) =  \pf{\vX , \vY}(x,y)\Indi_{z = 1}+ \pf{\vX}(x)\pf{\vY}(y) \Indi_{z = 0}.$$ 
Moreover, $\forall (x,y) \in \pOmegX \times \pOmegY$; we have by equation~\eqref{eq:SJlol1}:
$$\pf{\vW}(x, y) = \ratio \pf{\vX , \vY}(x,y) + (1 - \ratio) \pf{\vX}(x)\pf{\vY}(y).$$
So $(\vW, \vZ)$ satisfies the $\MDC(\ratio)$.

\vspace{0.5cm}
\noindent\STEPUIDFour\label{stepfour3}: \quad \begin{tabular}[t]{|l}
\begin{minipage}[t]{10cm}
\textbf{Concatenation.} Concatenate  the $3$ datasets  $\mDwzRes{\SampSizewzx}$, $\mDwzRes{\SampSizewzy}$ and $\mDwzRes{\SampSizewzxy}$
\end{minipage}
\end{tabular}\\

\vspace{0.5cm}
\noindent To conclude our proof, note that :\\

$\bullet$  Since $\mDxyRes{2\SampSizewzxy}$, $(\mDxyRes{\SampSizewzx},\mDxRes{\SampSizewzx})$ and $(\mDxyRes{\SampSizewzy}, \mDyRes{\SampSizewzy})$ are independent, the datasets $\mDwzRes{\SampSizewzx}$, $\mDwzRes{\SampSizewzy}$ and $\mDwzRes{\SampSizewzxy}$ are independent by construction.\\

$\bullet$ Moreover, since $\mDwzRes{\SampSizewzx}$, $\mDwzRes{\SampSizewzy}$ and $\mDwzRes{\SampSizewzxy}$ are composed by  $\pIID$ random variables following the law $\pf{\vW, \vZ}$, the sample of size $\SampSizewz = \SampSizewzxy+ \SampSizewzx + \SampSizewzy$,
$$\mDwz{\SampSizewz} =\mDwzRes{\SampSizewzx} \cup \mDwzRes{\SampSizewzy}\cup \mDwzRes{\SampSizewzxy} $$
is composed by  $\pIID$ random variables following the law $\pf{\vW, \vZ}$.

\begin{minipage}{4.3cm}
\begin{mdframed}
\vspace{-0.1cm}
\!\!\!\textbf{Construction of } $\mDwzRes{\SampSizewzxy}$.
\vspace{-0.4cm}
\end{mdframed}
\end{minipage}\\
\vspace{0.3cm}
If $\SampSizewzxy = 0$, then $\mDwzRes{\SampSizewzxy} = \emptyset$, otherwise we construct $\mDwzRes{\SampSizewzxy}$ from the initial dataset $\mDxRes{\SampSizewzx}$ as described in Construction~\ref{const:DwzNiid} (see the proof of Theorem~\ref{pro:ConsIID}).\\ 
\\
\begin{minipage}{4.3cm}
\begin{mdframed}
\vspace{-0.1cm}
\!\!\!\textbf{Construction of } $\mDwzRes{\SampSizewzy}$.
\vspace{-0.4cm}
\end{mdframed}
\end{minipage}\vspace{0.3cm}\\
First note that if $\SampSizewzy = 0$, then $\mDwzRes{\SampSizewzy}= \emptyset$. If $\SampSizewzy \neq 0$, we proceed as follows:\\

\begin{tabular}[t]{|c}\hspace{-0.5cm}
\begin{minipage}{12cm}
\begin{enumerate}[label=(\roman*)]
    \item Sample $\SampSizewzy$ independent observations $(\vZ_i)_{i = 1, \cdots, \SampSizewzy}$ according to a Bernoulli law of parameter $\ratio \in (0,1)$.
    \item For all $i = 1, \cdots, \SampSizewzy$; $\forall (\vX_i, \vY_i) \in \mDxyRes{\SampSizewzy}$ and $\forall \vYtild_i \in \mDyRes{\SampSizewzy}$, set $$\vW_i = (\vX_i, \vY_{i} \Indi_{\vZ_{i} = 1} + \vYtild_{i} \Indi_{\vZ_{ i} = 0}).$$
\end{enumerate}
\end{minipage}
\end{tabular}\\
\\
Note that by construction, $(\vW_i, \vZ_{i})_{i=1}^{\SampSizewzy}$ are $\pIID$ and follow $\pf{\vW,\vZ}$. Moreover, following the same arguments as in proof of Theorem~\ref{pro:ConsIID}, the $\{ (\vW_i, \vZ_{i})\}_{i=1}^{\SampSizewzy}$ satisfy the $\MDC(\ratio)$.\\
\\
\noindent
\begin{minipage}{4.3cm}
\begin{mdframed}
\vspace{-0.1cm}
\!\!\!\textbf{Construction of } $\mDwzRes{\SampSizewzx}$.
\vspace{-0.4cm}
\end{mdframed}
\end{minipage}\\
\vspace{0.3cm}
First note that if $\SampSizewzx = 0$, then $\mDwzRes{\SampSizewzx}= \emptyset$. If $\SampSizewzx \neq 0$, we consider the datasets $\mDxyRes{\SampSizewzx}$ and $\mDxRes{\SampSizewzx}$, and proceed as follows:\\

First construct a random vector $(\vW,\vZ)$ satisfying the $\MDC(\ratio)$ with $\ratio \in \pOmegr$ following a similar reasoning as in proof of Theorem~\ref{pro:ConsIID}: 
\begin{itemize}
    \item Let $(\vX,\vY)$ be a random variable admitting $\pf{\vX,\vY}$ as probability density function.
    \item Let $\vZ$ be a random variable following a Bernoulli of parameter $\ratio \in (0,1)$.
    \item Let $\vXtild$ be a random variable independent of $(\vX,\vY)$ following the law $\pf{\vX}$.
    \item Set $\vW = (\vX_{i} \Indi_{\vZ_{i} = 1} + \vXtild_{i} \Indi_{\vZ_{i} = 0}, \vY_i)$.
\end{itemize}
Then, $\forall (x, \widetilde{x},y) \in \pOmegX \times \pOmegY \times \pOmegY$ we have $\forall z \in \{0,1\}$,
$$
\pg{\vW \pCond \vZ = z}(x, \widetilde{x},y) = \left\{
\begin{array}{ll}
 \pf{\vX , \vY}(x,y) \text{ if }z = 1 \\
 \pf{\vX}(\widetilde{x})\pf{\vY}(y) \text{ if }z = 0
\end{array}
\right.$$\\
Then $\forall (x, y, z) \in \pOmegX \times \pOmegY \times \{0,1\}$, it comes $$\pf{\vW \pCond \vZ = z}(x, y) = \pg{\vW \pCond \vZ = z}(x, x,y) =  \pf{\vX , \vY}(x,y)\Indi_{z = 1}+ \pf{\vX}(x)\pf{\vY}(y) \Indi_{z = 0}.$$ 
Moreover, $\forall (x,y) \in \pOmegX \times \pOmegY$; we have by equation~\eqref{eq:SJlol1}:
$$\pf{\vW}(x, y) = \ratio \pf{\vX , \vY}(x,y) + (1 - \ratio) \pf{\vX}(x)\pf{\vY}(y).$$
So $(\vW, \vZ)$ satisfies the $\MDC(\ratio)$.\\

\begin{tabular}[t]{|c}\hspace{-0.5cm}
\begin{minipage}{12cm}
\begin{enumerate}[label=(\roman*)]
\item Sample $\SampSizewzx$ independent observations $(\vZ_i)_{i = 1, \cdots, \SampSizewzx}$ according to a Bernoulli law of parameter $\ratio \in (0,1)$.
\item For all $i = 1, \cdots, \SampSizewzx$; $\forall (\vX_i, \vY_i) \in \mDxyRes{\SampSizewzx}$ and $\forall \vXtild_i \in \mDxRes{\SampSizewzx}$, set $$\vW_i= (\vX_i \Indi_{z = 1}+ \vXtild_i \Indi_{z = 0}, \vY_i)$$.
\end{enumerate}
\end{minipage}
\end{tabular}\\
Note that, by construction $\mDwzRes{\SampSizewzx} = (\vW_i, \vZ_{i})_{i=1}^{\SampSizewzx}$ are $\pIID$ and follow $\pf{\vW,\vZ}$.\\
\\
\noindent
\begin{minipage}{4.3cm}
\begin{mdframed}
\vspace{-0.1cm}
\!\!\!\textbf{Concatenation.}
%\vspace{-0.1cm}
\end{mdframed}
\end{minipage}\\
\vspace{0.1cm}
Now to conclude our proof, note that $\mDwzRes{\SampSizewzx}$, $\mDwzRes{\SampSizewzy}$ and $\mDwzRes{\SampSizewzxy}$ are independent by construction, since $\mDxyRes{2\SampSizewzxy}$, $(\mDxyRes{\SampSizewzx},\mDxRes{\SampSizewzx})$ and $(\mDxyRes{\SampSizewzy}, \mDyRes{\SampSizewzy})$ are independent. Moreover, since $\mDwzRes{\SampSizewzx}$, $\mDwzRes{\SampSizewzy}$ and $\mDwzRes{\SampSizewzxy}$ samples are $\pIID$ random variables following the law $\pf{\vW, \vZ}$ we have $\mDwz{\SampSizewz} =\mDwzRes{\SampSizewzx} \cup \mDwzRes{\SampSizewzy}\cup \mDwzRes{\SampSizewzxy} $ a training set of $\SampSizewz = \SampSizewzxy+ \SampSizewzx + \SampSizewzy$ $\pIID$ samples following  the law $\pf{\vW, \vZ}$.

%%%%%%%%%%%%%%%%%%%%%%%%%%%%%%%%%%%%%%%%%%%%%
\subsection{Proof Theorem~\ref{pro:ConsUIDAM} and i.d. Construction in the Additional data setting}
 \label{Sec:ProofConsUIDAM}
 %%%%%%%%%%%%%%%%%%%%%%%%%%%%%%%%%%%%%%%%%%%%%

 Let $n_J$ and $n_M$ two integers such that $1 \leq n_J \leq \SampSizexy$ and $1 \leq n_M \leq (\SampSizexy + \SampSizex)(\SampSizexy + \SampSizey) - \SampSizexy$. 
 
 \vspace{0.5cm}
 %%%%%%%%%%%%%%%%%%%Construction 4%%%%%%%%%%%%%%%%%%%%%%%%%%%%%%%%%%%%%%%
\noindent
\begin{minipage}{8.5cm}
\begin{mdframed}
\vspace{-0.1cm}
\begin{const}
\label{const:DwzNidAD}
\big[$\pUID$ $\mDwz{\SampSizewz}$ Additional data\big]
\end{const}
\vspace{-0.1cm}
\end{mdframed}
\end{minipage}
\vspace{0.5cm}

\noindent The construction is the same as in the proof of Theorem~\ref{pro:ConsUID}, except we replace \STEPUIDOne in Construction~\ref{const:DwzNid} with \STEPUIDOneb detailed below.

\vspace{0.5cm}
\noindent\STEPUIDOneb\label{stepone4}: \quad \begin{tabular}[t]{|l}
\begin{minipage}[t]{10cm}
We first generate $(\SampSizexy + \SampSizex)(\SampSizexy + \SampSizey)$ observations $\{(\vW_i,\vZ_i)\}_{i = 1, \cdots, (\SampSizexy + \SampSizex)(\SampSizexy + \SampSizey)}$ by concatenating $4$ sets of samples denoted $\mS$, $\mMx$, $\mMy$ and $\mMxy$ of size $\SampSizexy^2$, $\SampSizexy \times \SampSizex$, $\SampSizexy  \times \SampSizey$ and $\SampSizex \times \SampSizey$ respectively.

\vspace{0.5cm}
\SUBSTEPA\label{steponeb} \quad \begin{tabular}[t]{||l}
\begin{minipage}[t]{10cm}
Generate the set of samples $\mS$ exactly like in \STEPUIDOne of the Construction~\ref{const:DwzNid}.
\end{minipage}
\end{tabular}
\vspace{0.5cm}

\SUBSTEPB \quad \begin{tabular}[t]{||l}
\begin{minipage}[t]{10cm}
 Construct the set of samples $\mMx = \{(\vW_i,\vZ_i)\}_{i = 1, \cdots, \SampSizexy\times \SampSizex }$ of size $\SampSizexy \times \SampSizex$ as follows:
$\forall j = 0, \cdots, \SampSizexy - 1, \, \forall k = 1, \cdots, \SampSizex$,
$$
\left\{
\begin{array}{ll}
\vW_{ j \SampSizex + k} =  (\vXtild_{k}, \vY_{\SampSizexy^2 + j + 1}) \\
\vZ_{ j \SampSizex + k} = 0
\end{array}
\right.
$$
\end{minipage}
\end{tabular}
%\vspace{0.5cm}

\end{minipage}
\end{tabular}\\

\noindent Since $\mDx$ and $\mDxy$ are independent and their respective elements are $\pIID$, we have $\forall i =  1 , \cdots , \SampSizexy^2 $, $\pf{\vW_i} \equiv \pf{\vX}\pf{\vY}$.

\vspace{0.5cm}
\hspace{1.8cm} \quad \begin{tabular}[t]{|l}
\begin{minipage}[t]{10cm}
\SUBSTEPC\quad \begin{tabular}[t]{||l}
\begin{minipage}[t]{10cm}
 Construct the set of samples $\mMy = \{(\vW_i,\vZ_i)\}_{i =  1, \cdots, \SampSizexy\times \SampSizey}$ of size $\SampSizexy \times \SampSizey$ as follows: $\forall j = 0, \cdots, \SampSizexy - 1, \, \forall k = 1, \cdots, \SampSizey$,
$$
\left\{
\begin{array}{ll}
\vW_{ j \SampSizey + k} =  (\vX_{ j + 1}, \vYtild_{k}) \\
\vZ_{ j \SampSizey + k} = 0
\end{array}
\right.
$$
\end{minipage}
\end{tabular}
\end{minipage}
\end{tabular}\\

\noindent Since $\mDy$ and $\mDxy$ are independent and their respective elements are $\pIID$, we have $\forall i =  1, \cdots , \SampSizexy\times \SampSizey$, $\pf{\vW_i} \equiv \pf{\vX}\pf{\vY}$.

\vspace{0.5cm}
\hspace{1.8cm} \quad \begin{tabular}[t]{|l}
\begin{minipage}[t]{10cm}
\SUBSTEPD\quad \begin{tabular}[t]{||l}
\begin{minipage}[t]{10cm}
 Construct the set of samples $\mMxy =\{(\vW_i,\vZ_i)\}_{i = 1, \cdots, \SampSizex \SampSizey }$ of size $\SampSizex \times \SampSizey$ as follows: $\forall j = 0, \cdots, \SampSizex - 1, \, \forall k = 1, \cdots, \SampSizey$,
$$
\left\{
\begin{array}{ll}
\vW_{ j \SampSizey + k} =  (\vXtild_{j + 1}, \vYtild_{k}) \\
\vZ_{ j \SampSizey + k} = 0
\end{array}
\right.
$$
\end{minipage}
\end{tabular}
\end{minipage}
\end{tabular}\\

\noindent Since $\mDx$ and $\mDy$ are independent and their respective elements are $\pIID$, we have $\forall i = 1, \cdots, \SampSizex \SampSizey$, $\pf{\vW_i} \equiv \pf{\vX}\pf{\vY}$.

\vspace{0.5cm}
\hspace{1.8cm} \quad \begin{tabular}[t]{|l}
\begin{minipage}[t]{10cm}
\SUBSTEPE\quad \begin{tabular}[t]{||l}
\begin{minipage}[t]{10cm}
Finally, concatenate $\mS$, $\mMx$, $\mMy$ and $\mMxy$ which ends the \STEPUIDOneb.
\end{minipage}
\end{tabular}
\end{minipage}
\end{tabular}\\

\vspace{0.5cm}
\noindent\STEPUIDNext\label{stepfour31}: \quad \begin{tabular}[t]{|l}
\begin{minipage}[t]{10cm}
Next, do \STEPUIDTwo, \STEPUIDThree, \STEPUIDFour and  \STEPUIDFive of the Construction~\ref{const:DwzNid}.
\end{minipage}
\end{tabular}

\vspace{0.5cm}

\noindent Using the same arguments as in the proof of Theorem~\ref{pro:ConsUID},  $\mDwz{\SampSizewz}$ satisfies the $\MDC(\ratio)$.
Since $\forall i = \SampSizexy^2 +1, \cdots, (\SampSizex + \SampSizexy) (\SampSizey + \SampSizexy)$, we have $\pf{\vW_i} \equiv \pf{\vX}\pf{\vY}$ and $\vZ_i = 0$. Therefore $\forall i = 1, \cdots , (\SampSizex + \SampSizexy) (\SampSizey + \SampSizexy)$, $\pf{\vW_i} \equiv \pf{\vX,\vY}  \Indi_{\vZ_i = 1} + \pf{\vX}\pf{\vY}  \Indi_{\vZ_i = 0}$.

%%%%%%%%%%%%%%%%%%%%%%%%%%%%%%%%%%%%%%%%%%%%%%%%
\subsection{Proof Theorem~\ref{pro:ConsUIDMT} and Construction in the non-independent case}
\label{Sec:ProofConsUIDMT}
%%%%%%%%%%%%%%%%%%%%%%%%%%%%%%%%%%%%%%%%%%%%%%%%

Let $n_J$ and $n_M$ two integers such that $1 \leq n_J \leq \SampSizexy \times \SampSizeM$ and $1 \leq n_M \leq \SampSizexy(\SampSizexy-1)\SampSizeM$. We construct the final dataset similary to Construction~\ref{const:DwzNid}, except we replace \STEPUIDOne with \STEPUIDOnet:\\

%%%%%%%%%%%%%%%%%construction 5%%%%%%%%%%%%%%%%%%%%%%%

\noindent
\begin{minipage}{8.5cm}
\begin{mdframed}
\vspace{-0.1cm}
\begin{const}
\label{const:DwzNidFram3}
\big[$\pUID$ $\mDwz{\SampSizewz}$ Framework~\ref{DATA3}\big]
\end{const}
\vspace{-0.1cm}
\end{mdframed}
\end{minipage}

\vspace{0.5cm}
\noindent\STEPUIDOnet\label{stepone5}: \quad \begin{tabular}[t]{|l}
\begin{minipage}[t]{10cm}
Construct $\SampSizexy^2\SampSizeM$ observations $\{(\vW_i,\vZ_i)\}_{i = 1, \cdots, \SampSizexy^2\SampSizeM}$ such that:
$$
\begin{array}{ll}
  \forall j = 0, \cdots, \SampSizexy - 1, \,& \forall k = 1, \cdots, \SampSizexy \, \forall l = 1, \cdots, \SampSizeM, \, \\
     & \left\{
\begin{array}{ll}
\vW_{(j \SampSizexy + k)\SampSizeM + l} =  (\vX_{j+1}, \vY^k_{l}) \\
\vZ_{(j \SampSizexy + k)\SampSizeM + l} = \Indi_{j+1 = k}
\end{array}
\right.
\end{array}
$$
\end{minipage}
\end{tabular}

\vspace{0.5cm}
\noindent\STEPUIDNext\label{stepfour32}: \quad \begin{tabular}[t]{|l}
\begin{minipage}[t]{10cm}
Next, do {}\STEPUIDTwo, \STEPUIDThree, \STEPUIDFour and  \STEPUIDFive of the Construction~\ref{const:DwzNid}.
\end{minipage}
\end{tabular}\\

\noindent Since  $\forall j = 0, \cdots, \SampSizexy - 1, \, \forall k = 1, \cdots, \SampSizexy, \,\, \forall l = 1, \cdots, \SampSizeM,$ 
$$
\vX_{j+1} \pIndep \vY^k_{l} \ iff \ j+1 \neq k,
$$
we can use the same arguments as in the proof of Theorem~\ref{pro:ConsUID} to show that $\forall i = 1, \cdots , \SampSizewz,$ the couple $(\vW_i, \vZ_i) \in \mDwz{\SampSizewz}$ satisfies the $\MDC\left(\frac{n_J}{\SampSizewz}\right)$.

%% file: Appendix.tex
\section{Appendix}
\subsection{Method to rescale estimated densities}
If we want our estimation to correspond to a proper probability density function, that is $\Integ{\pOmegY}{}\,\pfhat{\vY \pCond \vX = x}(y)dy \approx 1$, we can numerically estimate the integral using the trapezoidal rule on a grid of $m$ target values $\{y_i\}_{i = 1, \cdots, m}$ such that its values are evenly distributed between the quantiles $0.001$ and $0.999$ of $\pf{\vY}$ (which we can always estimate through the marginal density estimator) and then rescale the predicted value, doing as follows: \\\\
\noindent
\rescale \quad \begin{tabular}[http]{|l}
\begin{minipage}[t]{10cm}
    $\bullet$ Generate a grid of $m$ target values $\{y_i\}_{i = 1, \cdots, m}$. \\
    $\bullet$ Estimate $\{\pf{\vY \pCond \vX = x}(y_i)\}_{i = 1, \cdots, m}$.\\
    $\bullet$ Use the trapezoidal rule to estimate $\Integ{\pOmegY}{}\,\pfhat{\vY \pCond \vX = x}(y)dy$.\\
    $\bullet$ Divide the predicted value by the estimation of $\Integ{\pOmegY}{}\,\pfhat{\vY \pCond \vX = x}(y)dy$. 
\end{minipage}
\end{tabular}\\

.
\\\\
Note that this step must be repeated for each observation $x$ for which we want to estimate a conditional density function, but given a set observation $x$, we can reuse the computed integral for any number of target values. This technique will also be used for the other benchmarked methods which do not yield proper integrals by default.

\subsection{Python code for training set construction}
We provide extracts of the \packpython code used to construct $\mDwz{}$, the training set for the discriminator.
\begin{lstlisting}[language=Python, caption=Python code extract corresponding to Constructions 1 and 3.]
import numpy as np
from sklearn.model_selection import train_test_split as tts
from sklearn.utils import shuffle
...
n_obs, p = X.shape
n_extraobs, n_extrasamples = min(len(extraobs),n_obs), min(len(extrasamples),n_obs)
n_extra = min(max(n_extraobs, n_extrasamples),n_obs)
n_contrast = int((n_obs + n_extra )/2)
if n_extra == 0:
    X_joint, X_discarded, y_joint, y_marginal = tts(X, y, train_size = n_contrast, random_state = self._random_state)
    X_marginal = X_joint
elif n_extra == n_obs:
    if n_extraobs <= n_extrasamples:
        X_joint, X_marginal = X, X
        y_joint, y_marginal = y, extrasamples[:n_obs]
    else: 
        y_joint, y_marginal = y, y
        X_joint, X_marginal = X, extraobs[:n_obs]
elif n_extraobs <= n_extrasamples:
    X_joint, X_discarded, y_joint, y_marginal_partial = tts(X,y, train_size = n_contrast, random_state = self._random_state)
    X_marginal = X_joint
    y_marginal = np.concatenate([y_marginal_partial, extrasamples[:2 * n_contrast - n_obs]], axis= 0)
elif n_extraobs > n_extrasamples:
    X_joint, X_marginal_partial, y_joint, y_discarded = tts(X, y, train_size = n_contrast, random_state = self._random_state)
    y_marginal = y_joint
    X_marginal = np.concatenate([X_marginal_partial, extraobs[:2 * n_contrast - n_obs]], axis= 0)
z = self._random_state.binomial(1, self.contrast_ratio, size = (n_contrast))
X_joint, X_marginal, y_joint, y_marginal = X_joint[:n_contrast], X_marginal[:n_contrast], y_joint[:n_contrast].reshape((n_contrast,-1)), y_marginal[:n_contrast].reshape((n_contrast,-1))
W_X = X_joint * z.reshape((-1,1)) + X_marginal * (1 - z.reshape((-1,1)))
W_y = y_joint * z.reshape((-1,1)) + y_marginal * (1 - z.reshape((-1,1)))
W = self._build_W(W_X, W_y)
...
def _build_W(self, X, y):
    if len(X.shape)!= len(y.shape):
        y = y.reshape((len(X), -1))
    return np.concatenate([X, y], axis = -1)
\end{lstlisting}
\begin{lstlisting}[language=Python, caption=Python code extract corresponding to Constructions 2 and 4.]
import numpy as np
from sklearn.model_selection import train_test_split as tts
from sklearn.utils import shuffle
...
n_obs, p = X.shape
n_joint = n_obs
if self.contrast_ratio < 1. / n_obs: self.contrast_ratio = 1. / n_obs #does not consider the case where p < 1 / n  <==> n_J < n, as there is no practical reason to make this choice
n_marginal = int(n_joint / self.contrast_ratio - n_joint)
n_marginal_extra = min(len(extraobs) * len(extrasamples), n_marginal)
n_marginal_x = min(len(extraobs) * n_obs, n_marginal - n_marginal_extra)
n_marginal_y = min(n_obs * len(extrasamples), n_marginal - n_marginal_extra - n_marginal_x)
n_marginal_both = min(n_obs * (n_obs - 1), n_marginal - n_marginal_extra - n_marginal_y - n_marginal_x)

W_joint = self._build_W( X, y)

W_marginal = np.zeros((0, W_joint.shape[1]))
for n_added_marginal, X_added, y_added, remove_same_index in [(n_marginal_extra, extraobs, extrasamples, False),
                                           (n_marginal_x, extraobs, y, False),
                                           (n_marginal_y, X, extrasamples, False),
                                           (n_marginal_both, X, y, True)]:
    if n_added_marginal:
        W_marginal_added = self._shuffle_and_sample(X_added, y_added, n_added_marginal, remove_same_index = remove_same_index)
        W_marginal = np.concatenate([W_marginal, W_marginal_added], axis = 0)

W = np.concatenate([W_joint, W_marginal], axis = 0)
z = (np.arange(n_joint + n_marginal) < n_joint).astype(int)
shuffled_indexes = shuffle(np.arange(n_joint + n_marginal), random_state = self._random_state)
W, z = W[shuffled_indexes], z[shuffled_indexes]
#to insure we use the same distribution the discriminator has seen during training
true_contrast_ratio = z.mean() 
...
def _shuffle_and_sample(self, X, y, n_samples, remove_same_index = False):
    len_x, len_y = len(X), len(y)
    x_indexes = np.arange(len_x * len_y) % len_x
    y_indexes = np.arange(len_x * len_y) // len_x
    if remove_same_index:
        isnot_joint = x_indexes != y_indexes
        x_indexes, y_indexes = x_indexes[isnot_joint], y_indexes[isnot_joint]
    if n_samples < len(x_indexes):
        sampled_indexes = self._random_state.choice(np.arange(len(x_indexes)), size = n_samples, replace= False)
        x_indexes, y_indexes = x_indexes[sampled_indexes], y_indexes[sampled_indexes]
    return self._build_W( X[x_indexes], y[y_indexes])
\end{lstlisting}
\begin{lstlisting}[language=Python, caption=Python code extract corresponding to Construction 5.]
import numpy as np
from sklearn.model_selection import train_test_split as tts
from sklearn.utils import shuffle
...
n_obs, p = X.shape
n_joint = n_obs
#target values are stored in extrasamples
#extrasamples.shape = (n, m)
multi_sample_size = extrasamples.shape[-1]
if self.contrast_ratio < 1. / (n_obs * multi_sample_size) : self.contrast_ratio = 1. / (n_obs * multi_sample_size) #does not consider the case where r < 1 / n  <==> n_J < n, as there is no practical reason to make this choice
n_joint = n_obs * multi_sample_size
n_total = int(n_joint / self.contrast_ratio)
n_marginal = n_total - n_joint

X_joint = X.repeat(multi_sample_size,axis=0)
y_joint = extrasamples.reshape((n_obs * multi_sample_size,-1))
W_joint = self._build_W(X_joint, y_joint)

X_marginal_max = []
y_marginal_max = []
for i in range(n_obs):
    for j in range(n_obs):
        if i != j:
            X_marginal_max.append(X[i:i+1].repeat(multi_sample_size,axis=0))
            y_marginal_max.append(extrasamples[j:j+1].reshape((multi_sample_size,-1)))
X_marginal_max = np.concatenate(X_marginal_max, axis = 0)
y_marginal_max = np.concatenate(y_marginal_max, axis = 0)
X_marginal, X_discarded, y_marginal, y_discarded = tts(X_marginal_max, y_marginal_max, train_size = n_marginal, random_state = self._random_state)
W_marginal = self._build_W(X_marginal, y_marginal)

W = np.concatenate([W_joint, W_marginal], axis = 0)
z = (np.arange(n_joint + n_marginal) < n_joint).astype(int)
shuffled_indexes = shuffle(np.arange(n_joint + n_marginal), random_state = self._random_state)
W, z = W[shuffled_indexes], z[shuffled_indexes]
#to insure we use the same distribution the discriminator has seen during training
true_contrast_ratio = z.mean() 
\end{lstlisting}

\subsection{Exhaustive experimental results}
In this section, we present the exhaustive results corresponding to Tables~\ref{table:features}, ~\ref{table:timing} and ~\ref{table:real}.
\begin{table}[http]
\centering
\tiny
\begin{tabular}{lccccccccccc}
Empirical \pKL & \rot{\mcd:\MLP} & \rot{\mcd:\CAT} & \rot{\nnk} & \rot{\mdn} & \rot{\kmn} & \rot{\nf} & \rot{\lsc} & \rot{\rfcde} & \rot{\flex:\MLP} & \rot{\flex:\XGB} & \rot{\dcde} \\
\hline
$\DimInput = 3$ & 0.008 & 0.009 & 0.022 & 0.109 & 0.067 & 0.189 & 0.342 & 0.037 & 0.094 & 0.093 & 0.152 \\
$\DimInput = 10$ & 0.036 & 0.042 & 0.063 & 0.229 & 0.096 & 0.228 & 0.61 & 0.105 & 0.106 & 0.076 & 0.205 \\
$\DimInput = 30$ & 0.115 & 0.202 & 0.154 & 0.396 & 0.181 & 0.432 & 0.369 & 0.238 & 0.22 & 0.196 & 0.385 \\
$\DimInput = 100$ & 0.162 & 0.224 & 0.173 & 0.45 & 0.401 & 0.411 & 1.059 & 0.238 & 0.234 & 0.264 & 0.36 \\
$\DimInput = 300$ & 0.244 & 0.308 & 0.282 & 0.506 & 0.304 & 0.67 & 0.783 & 0.507 & 0.253 & 0.302 & 0.306 \\
\hline
\end{tabular}
\caption{Evaluation of the \pKL divergence values for various feature sizes $\DimInput$, on the \dBasic density model, with $\SampSizexy= 100$.}
\end{table}

\begin{table}[http]
\centering
\tiny
\begin{tabular}{lccccccccccc}
Time in sec. & \rot{\mcd:\MLP} & \rot{\mcd:\CAT} & \rot{\nnk} & \rot{\mdn} & \rot{\kmn} & \rot{\nf} & \rot{\lsc} & \rot{\rfcde} & \rot{\flex:\MLP} & \rot{\flex:\XGB} & \rot{\dcde} \\
\hline

$\SampSizexy=30$ & 3.133 & 0.46 & 0.01 & 35.46 & 47.92 & 50.74 & 12.01 & 0.044 & 0.018 & 9.574 & 1.943 \\
$\SampSizexy=100$ & 3.149 & 0.369 & 0.022 & 35.45 & 102.7 & 50.72 & 37.14 & 0.149 & 0.017 & 9.57 & 2.11 \\
$\SampSizexy=300$ & 3.132 & 0.446 & 0.04 & 35.65 & 150.6 & 50.62 & 100.5 & 0.448 & 0.017 & 13.19 & 3.225 \\
$\SampSizexy=1000$ & 3.296 & 0.689 & 0.043 & 35.73 & 143.6 & 51.12 & 167.9 & 0.952 & 0.02 & 25.24 & 6.388 \\
\hline
\end{tabular}
\caption{Training Time in seconds for various training set sizes $\SampSizexy$, on the \dBasic density model.}
\end{table}

\begin{table}[http]
\centering
\tiny
\begin{tabular}{lcccccccccc}
Empirical \pNLL & \rot{\mcd:\MLP} & \rot{\CAT} & \rot{\nnk} & \rot{\mdn} & \rot{\kmn} & \rot{\nf} & \rot{\lsc} & \rot{\flex:\MLP} & \rot{\flex:\XGB} & \rot{\dcde} \\
\hline
BostonHousing & -0.64 & -0.59 & -0.81 & -1.17 & -1.22 & -1.63 & -2.19 & -1.99 & -1.84 & -7.09 \\
Concrete & -0.86 & -1.02 & -1.23 & -2.02 & -2.30 & -2.13 & -4.00 & -2.26 & -1.25 & -10.1 \\
NCYTaxiDropoff:lon. & -1.30 & -1.28 & -1.68 & -2.51 & -2.51 & -2.85 & -3.90 & -2.05 & -2.17 & -11.2 \\
NCYTaxiDropoff:lat. & -1.31 & -1.31 & -1.44 & -2.51 & -2.45 & -2.96 & -4.72 & -1.67 & -6.18 & -9.35 \\
Power & -0.06 & -0.36 & -0.73 & -0.55 & -0.35 & -0.39 & -0.70 & -1.09 & -0.75 & -8.97 \\
Protein & -0.09 & -0.42 & -0.77 & -0.54 & -0.68 & -0.54 & -1.09 & -0.83 & -1.38 & -10.5 \\
WineRed & -0.89 & -0.89 & 3.486 & -1.27 & -0.90 & -2.43 & -6.25 & 1.062 & 0.965 & -10.1 \\
WineWhite & -1.18 & -1.13 & 2.99 & -2.24 & -1.7 & -4.18 & -5.41 & -0.73 & -0.63 & -13.2 \\
Yacht & 0.14 & 0.822 & -0.46 & 0.083 & 0.025 & 0.401 & -2.79 & -1.23 & 0.144 & -7.52 \\
teddy & -0.47 & -0.51 & -0.83 & -0.87 & -0.76 & -0.94 & -0.91 & -0.83 & -1.34 & -9.59 \\
toy dataset 1  & -0.99 & -0.47 & -0.63 & -0.40 & -0.35 & -0.71 & -0.70 & -0.88 & -1.46 & -6.10 \\
toy dataset 2 & -1.40 & -1.33 & -1.31 & -1.40 & -1.33 & -1.39 & -1.35 & -1.54 & -1.43 & -3.53 \\
\hline
\end{tabular}
\caption{Evaluation of the negative log-likelihood (\pNLL) for $12$ datasets.}
\end{table}